\documentclass{article}

\usepackage{PRIMEarxiv}

\usepackage[utf8]{inputenc} 
\usepackage[T1]{fontenc}    
\usepackage{hyperref}       
\usepackage{url}            
\usepackage{booktabs}       
\usepackage{amsfonts}       
\usepackage{nicefrac}       
\usepackage{microtype}      
\usepackage{lipsum}
\usepackage{fancyhdr}       
\usepackage{graphicx}       
\graphicspath{{media/}}     

\usepackage{subfigure}
\usepackage{xcolor}
\usepackage{float}

\usepackage{amsmath}
\usepackage{amssymb}
\usepackage{mathtools}
\usepackage{amsthm}

\usepackage[capitalize,noabbrev]{cleveref}

\theoremstyle{plain}
\newtheorem{theorem}{Theorem}[section]

\theoremstyle{definition}
\newtheorem{definition}[theorem]{Definition}

\theoremstyle{remark}

\DeclareMathOperator*{\minimize}{\textrm{minimize}}

\newcommand{\R}{\mathbb{R}}
\newcommand{\expect}{\mathop{\mathbb{E}}}
\newcommand{\observer}{\underset{\lambda,\vartheta}{\mathcal{H}}}
\newcommand{\model}{\underset{\lambda,\vartheta}{\mathcal{M}}}

\newcommand{\twopartdef}[4]{
	\left\{
		\begin{array}{ll}
			#1 & \mbox{if } #2 \\
			#3 & \mbox{if } #4
		\end{array}
	\right.
}

\newcommand{\threepartdef}[6]
{
	\left\{
		\begin{array}{lll}
			#1 & \mbox{if } #2 \\
			#3 & \mbox{if } #4 \\
			#5 & \mbox{if } #6
		\end{array}
	\right.
}

\title{Low-Resource White-Box Semantic Segmentation of Supporting Towers on 3D Point Clouds via Signature Shape Identification
}

\author{
  Diogo Lavado, Cláudia Soares \\
  NOVA School of Science and Technology \\
  Lisbon\\
  \texttt{\{d.lavado,claudia.soares\}@fct.unl.pt} 
   \And
  Alessandra Micheletti, Giovanni Bocchi \\
  University of Milan,\\
  Milan\\
  \texttt{\{alessandra.micheletti,giovanni.bocchi1\}@unimi.it} \\
  \And
  Alex Coronati, Manuel Silva\\
  EDP NEW,\\
  Lisbon,\\
  \texttt{\{alex.coronati,manuelpio.silva\}@edp.pt} \\
  \AND
  Patrizio Frosini,\\
  University of Bologna,\\
  Bologna,\\
  \texttt{patrizio.frosini@unibo.it}
}

\begin{document}
\maketitle

\begin{abstract}

Research in 3D semantic segmentation has been increasing performance metrics, like the IoU, by scaling model complexity and computational resources, leaving behind researchers and practitioners that (1) cannot access the necessary resources and (2) do need transparency on the model decision mechanisms.
%
%
In this paper, we propose SCENE-Net, a low-resource white-box model for 3D point cloud semantic segmentation. SCENE-Net identifies signature shapes on the point cloud via group equivariant non-expansive operators (GENEOs), providing intrinsic geometric interpretability.
%
Our training time on a laptop is 85~min, and our inference time is 20~ms. SCENE-Net has 11 trainable geometrical parameters,
and requires fewer data than black-box models. SCENE--Net offers robustness to noisy labeling and data imbalance and has comparable IoU to state-of-the-art methods. 
With this paper, we release a 40~000 Km labeled dataset of rural terrain point clouds and our code implementation.

\end{abstract}

\section{Introduction}
\label{sec:intro}
%

\begin{figure}[t]
\centering
\subfigure[TS40K Sample]{\label{fig:intro_input}\includegraphics[width=.29\columnwidth]{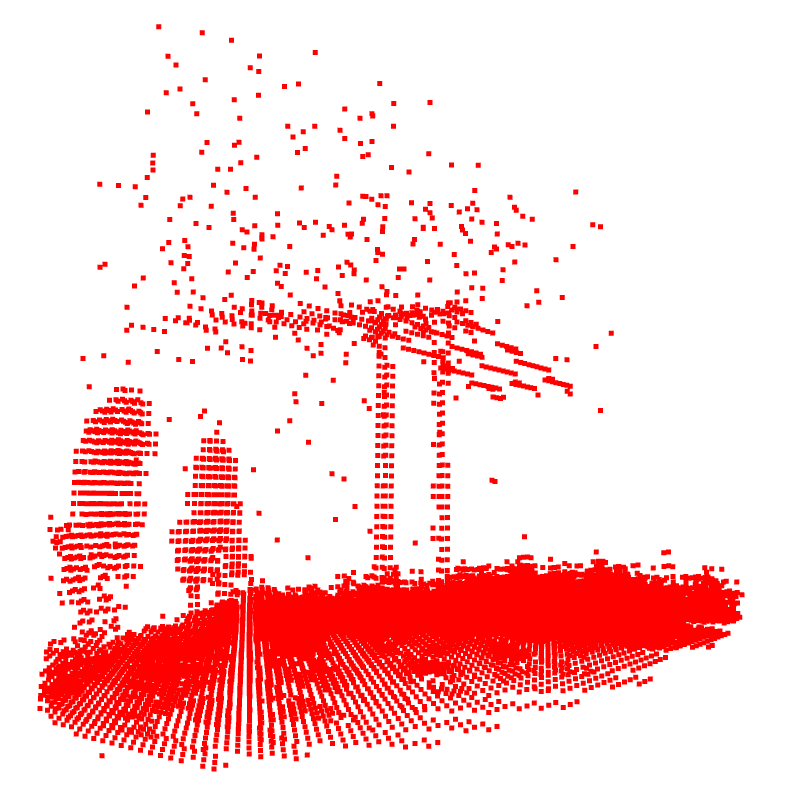}}
\subfigure[SCENE-Net]{\label{fig:intro_gnet}\includegraphics[width=.37\columnwidth]{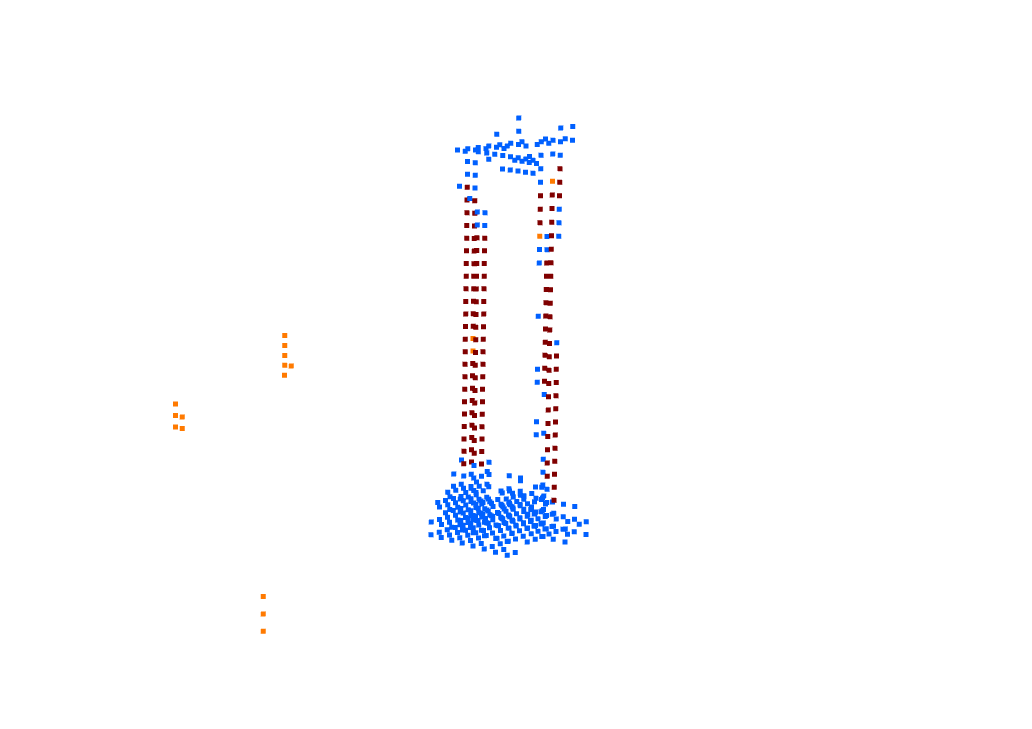}}
\subfigure[Baseline CNN]{\label{fig:intro_cnn}\includegraphics[width=.32\columnwidth]{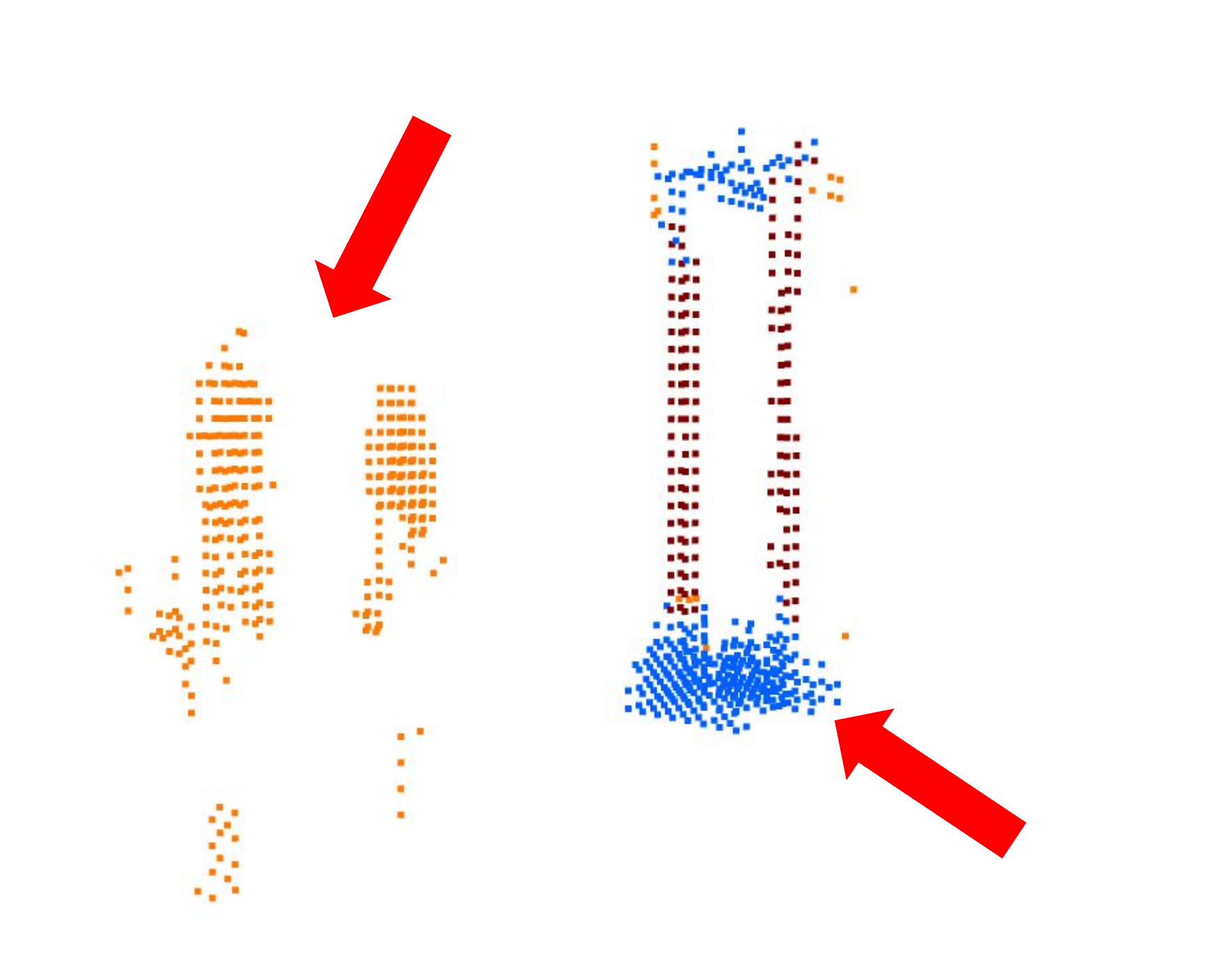}}
\subfigure
{\centering\includegraphics[width=.5\columnwidth]{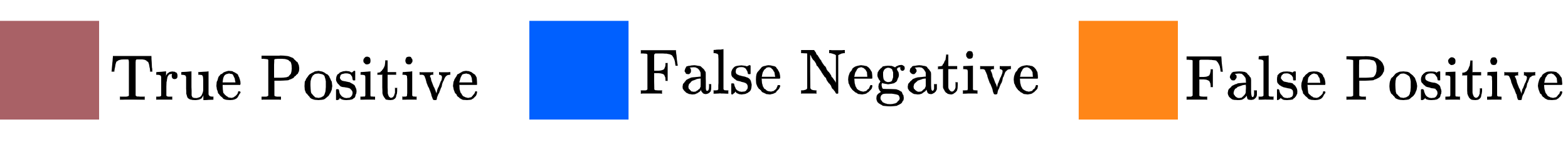}}
\caption{Signature shapes for power line supporting tower detection.
For our TS40K sample shown in (a), SCENE-Net accurately detects the body of the tower (b), while a comparable CNN has a large false positive area in the vegetation (c). Our model is interpretable with 11 trainable geometric parameters whereas the CNN has a total of 2190 parameters. The ground and power lines are mislabeled in the ground truth.}
\label{fig:intro_fig}
\end{figure}

%

Powerful Machine Learning (ML) algorithms applied to critical applications, such as autonomous driving or environmental protection, highlight the importance of (1) ease of implementation for non-tech organizations entailing data efficiency and general-purpose hardware, and (2) transparent models regarding their decision-making process, thus ensuring a responsible deployment~\cite{lipton2018mythos,guidotti2018survey, doshi2017towards}. 
%
%
%
Most methods in Explainable AI (XAI) provide \textit{post hoc} explanations to black-box models (i.e., algorithms unintelligible to humans). However, these are often limited in terms of their model fidelity~\cite{lipton2018mythos,rudin2019stop}, that is, they provide explanations for the predictions of the underlying model (e.g., heatmaps~\cite{Chefer_2021_CVPR,voita2019analyzing} and input masks~\cite{ribeiro2016should,fong2017interpretable}), instead of providing a mechanistic understanding of its inner-workings.
%
%
%
Conversely, intrinsic interpretability methods (i.e., white-box models) provide an
understanding of their decisions through their architecture
and parameters~\cite{rudin2019stop}.
 Transparency is achieved by enforcing constraints that reflect domain knowledge and simple designs~\cite{lipton2018mythos,chen2020concept}, which can result in a loss in performance when compared to complex black-boxes.

We propose a novel white-box model, \textbf{SCENE-Net}, that provides intrinsic geometric interpretability by leveraging on group equivariant non-expansive operators (GENEOs)~\cite{GENEO19,GENEO21}. Unlike traditional interpretable models, GENEOs are complex observers parameterized with meaningful geometric features.
In our case, task dependency comes as a collaboration of Machine Learning and electrical utility teams to transparently segment
power line supporting towers on 3D point clouds to inspect extensive power grids automatically.
Electrical grid operators have the critical job of assessing the risk of contact between the power grid and its environment to prevent failures and forest fires. These grids spread over countries and even continents, thus making careful inspection an important and challenging problem. 
Often, this task is based on LiDAR large-scale point clouds with high-point density, no sparsity, and no object occlusion. However, the captured point clouds are quite extensive and mostly composed of rural areas. 
These data are different from large urban datasets for autonomous driving~\cite{Semantic3D,SemKITTI} due to the point of view, point density, occlusion, and extension.
To bootstrap this work, we created a labeled dataset of 40~000~Km of rural and forest terrain, and the \textbf{T}ransmission \textbf{S}ystem, named \textbf{TS40K}. 
These point clouds show noisy labels and class imbalance (see Appendix~\ref{sec:ts40k} for details), and our SCENE-Net is robust to labeling noise as it encodes the geometric properties we need to detect.

Moreover, practitioners in high-risk tasks, such as autonomous driving and power grid inspection, are often limited in terms of resources, namely computational power and available data, to train and deploy state-of-the-art models~\cite{muhammad2020deep,alzubaidi2021review}.
This clashes with the current trend in DL to scale up models in both complexity and needed resources in order to maximize performance, for example, state-of-the-art 3D semantic segmentation models~\cite{thomas2019kpconv,tang2020searching,xu2021rpvnet,yan2021sparse,yan20222dpass} follow this trend.
Our model, SCENE-Net, maintains a simple design conventional to white-boxes (it is composed of 11 trainable parameters) that allows for resource-efficient training while taking advantage of powerful Deep Learning (DL) strategies, such as convolutions.
By assessing our model on the SemanticKITTI benchmark~\cite{SemKITTI}, we show that SCENE-Net achieves performance on-par with state-of-the-art methods in pole segmentation.

Our main contributions are:
\begin{itemize}
   
    \item SCENE-Net is the first white-box model for 3D semantic segmentation on large-scale landscapes, including non-urban environments (Section~\ref{sec:method});
   
    \item The architecture of SCENE-Net has fewer trainable parameters than traditional methods and is resource-efficient in both data and computational requirements (Section~\ref{sec:results-time-efficiency});

    \item Empirically, SCENE-Net is intrinsically and posthoc interpretable and robust under noisy labels, with au par IoU (Section~\ref{sec:results-interpretability});
    
    \item We present TS40K, a new 3D point cloud dataset covering 40 000 Km of non-urban terrain, with more than 9000 million 3D points (details in Appendix~\ref{sec:ts40k});
\end{itemize}

\section{Related Work}
\label{sec:related-work}

\paragraph*{Point Cloud Semantic Segmentation.}
Processing point clouds is a challenging task due to their unstructured nature and invariance to permutations. 
%
Voxel-based strategies endow point clouds with structure in order to apply 3D convolutions~\cite{long2015fully,rethage2018fully,tchapmi2017segcloud}. However, memory footprint is too large for high-resolution voxel grids, while low resolution entails information loss. 
Subsequent methods try to answer these issues by employing sparse convolutions~\cite{graham20183d,su2018splatnet} and octree-based CNNs~\cite{wang2017cnn,le2018pointgrid}.
Point-based models take point clouds directly as input.
The work of PointNet~\cite{qi2017pointnet} and PointNet++~\cite{qi2017pointnet++} inspired the use of point sub-sampling strategies with feature aggregation techniques to learn local features on each sub-point~\cite{hu2020randla,xu2020geometry}.
Convolution-based methods~\cite{hua2018pointwise,li2018pointcnn,wu2019pointconv,thomas2019kpconv,Cylinder3D} demonstrate good performance on 3D semantic segmentation benchmarks, such as \textit{SemanticKITTI}~\cite{SemKITTI} and SensatUrban~\cite{SensatUrban}.
Following this strategy, recent methods exploit multi-representation fusion, i.e, they combine different mediums (voxel grids, raw point clouds, and projection images) to boost feature retrieval~\cite{AF2S3Net,tang2020searching,xu2021rpvnet,yan20222dpass} and achieve top performance on the above benchmarks.
While voxel-based methods are computationally expensive due to 3D convolutions on high-resolution voxel grids, point-based strategies have to use costly neighbor searching to extract local information.
We propose a voxel-based architecture that is time-efficient with high-resolution voxel grids, with shapes of $64^3$ and $128^3$.
Moreover, learning from imbalanced and noisy data is still a challenging task in point cloud segmentation~\cite{guo2020deep},
SCENE-Net is interpretable and robust to these conditions.

\paragraph*{Explainable Machine Learning.}
%
%
%

Explainability is a crucial aspect of ML methods in high-stakes tasks such as autonomous driving~\cite{lipton2018mythos,guidotti2018survey,doshi2017towards}.
Two main approaches have been proposed in the literature: \textit{post hoc} explainability, and intrinsic interpretability.
\textit{Post hoc} methods, such as LIME~\cite{ribeiro2016should}, meaningful perturbations~\cite{fong2017interpretable}, anchors~\cite{ribeiro2018anchors}, and ontologies~\cite{leite21}, are applied to trained black-box models and provide instance-based explanations that correlate model predictions to the given input. 
These methods are model-agnostic, and thus more flexible, but they often lack mechanistic cause-effect relations and have a limited understanding of feature importance~\cite{rudin2019stop}.
For instance, a dog image and random noise may generate similar importance heatmaps for the same class with the LIME method~\cite{rudin2019stop}.
Moreover, they introduce computational overhead, which may limit their application in real-world scenarios with complex black-box models, such as in the 3D semantic segmentation task.
%
%
%
%
%

In contrast, intrinsic interpretability methods provide an understanding of their decisions through their architecture and parameters~\cite{rudin2019stop}. Decision trees and linear regression are examples of white-box models. However, transparency is usually achieved by imposing domain constraints and simple designs, which implies limited performance compared to deep neural networks.
Recent advances in interpretable techniques, such as concept whitening~\cite{chen2020concept} and interpretable CNNs~\cite{Zhang_2018_CVPR}, have shown that interpretability does not have to imply performance loss. 
However, these methods provide evidential interpretability, that is, they offer intrinsic explanations to model predictions that are still linked to human interpretations and may imply an evidential correlation, but not causation.

We propose a white-box model, SCENE-Net, with intrinsic geometric interpretability that is not subject to human interpretation.
SCENE-Net analyzes the input 3D space according to prior knowledge of the geometry of objects of interest, which is encoded in functional observers and whose parameters are fine-tuned during training. These observers encode high-level geometrical concepts.
Thus, our predictions exhibit direct mechanistic cause-effect w.r.t. the learned observers. 
SCENE-Net maintains a simple model design in high-level mathematical operations while taking advantage of DL complex convolutional kernels. 


\section{Group Equivariant Non-Expansive Operators (GENEOs).} 

GENEOs are the building blocks of a mathematical framework~\cite{GENEO19} that formally describes machine learning agents as a set of operators acting on the input data. 
These operators provide a measure of the world, just as CNN kernels learn essential features to, for instance, recognize objects. 
Such agents can be thought of as observers that analyze data. They transform it into higher-level representations while respecting a set of properties (i.e., a group of transformations).
%
An appropriate observer transforms data in such a way that respects the right group of transformations, that is, it commutes with these transformations. Formally, we say that the observer is \textit{equivariant} with respect to a group of transformations. 
The framework takes advantage of topological data analysis (TDA) to describe data as topological spaces. 
Specifically, a set of data $X$ is represented by a topological space $\Phi$ with admissible functions $\varphi \colon X\to \R^3$. $\Phi$ can be thought of as a set of admissible measurements that we can perform on the measurement space $X$. For example, images can be seen as functions assigning RGB values to pixels. 
This not only provides uniformity to the framework but also allows us to shift our attention from raw data to the space of measurements that characterizes it.
Now that the input data is well represented, let us introduce how the framework defines prior knowledge. 
Data properties are defined through maps from $X$ to $X$ that are $\Phi$-preserving homeomorphisms. 
That is, the composition of functions in $\Phi$ with such homeomorphisms produces functions that still belong to $\Phi$.
Therefore, we can define a group $G$ of $\Phi$-preserving homeomorphisms, representing a group of transformations on the input data for which we require equivariance to be respected.
In other words, $G$ is the group of properties that we chose to enforce equivariance w.r.t. the geometry in the original data. 
It is through $G$ that we embed prior knowledge into a GENEO model. Following the previous example, planar translations can define a subgroup of $G$.

Let us consider the notion of a \textit{perception pair} $(\Phi, G)$: it is composed of all admissible measurements $\Phi$ and a subgroup of $\Phi$-preserving homeomorphisms $G$.
\begin{definition}[Group Equivariant Non-Expansive Operator (GENEO)]
Consider two perception pairs $(\Phi, G)$ and $(\Psi, H)$ and a homomorphism $T\colon G \to H$. A map $F\colon \Phi \to \Psi$ is a group equivariant non-expansive operator if it exhibits equivariance:
\begin{equation}
    \forall \varphi \in \Phi, \forall g \in G, F(\varphi  \circ g) = F(\varphi) \circ T(g)
\end{equation}

and is non-expansive:
\begin{align}
\begin{split}
    \forall \varphi_1, \varphi_2 \in \Phi,
    \Vert F(\varphi_1) - F(\varphi_2) \Vert_{\infty} &\leq \Vert \varphi_1 - \varphi_2 \Vert_{\infty}
\end{split}
\end{align}
\end{definition}

Non-expansivity and convexity are essential for the applicability of GENEOs in a machine-learning context.
When the spaces $\Phi$ and $\Psi$ are compact, non-expansivity guarantees that the space of all GENEOs $\mathcal{F}$ is compact as well. 
Compactness ensures that any operator can be approximated by a finite set of operators sampled in the same space. 
%
Moreover, by assuming that $\Psi$ is convex, \cite{GENEO19} proves that $\mathcal{F}$ is also convex. 
Convexity guarantees that the convex combination of GENEOs is also a GENEO. Therefore, these results prove that any GENEO can be efficiently approximated by a certain number of other GENEOs in the same space.
%

In addition to drastically reducing the number of parameters in the modeling of the considered problems and making their solution more transparent, we underline that the use of GENEOs makes available various theoretical results that allow us to take advantage of a new mathematical theory of knowledge engineering.
We stress that, besides the cited compactness and convexity theorems,
algebraic methods concerning the construction of GENEOs are already
available~\cite{bocchi2022geneonet,conti2022construction,botteghi2020finite}


\section{SCENE-Net: Signature geometriC Equivariant Non-Expansive operator Network}
\label{sec:method}
In this section, we introduce the overall architecture of \textbf{SCENE-Net}. Next, we define the geometrical properties that describe power line supporting towers. Lastly, we detail the loss function used to train the observer.
\begin{figure}[t]
    \centering
    \includegraphics[width=.9\columnwidth]{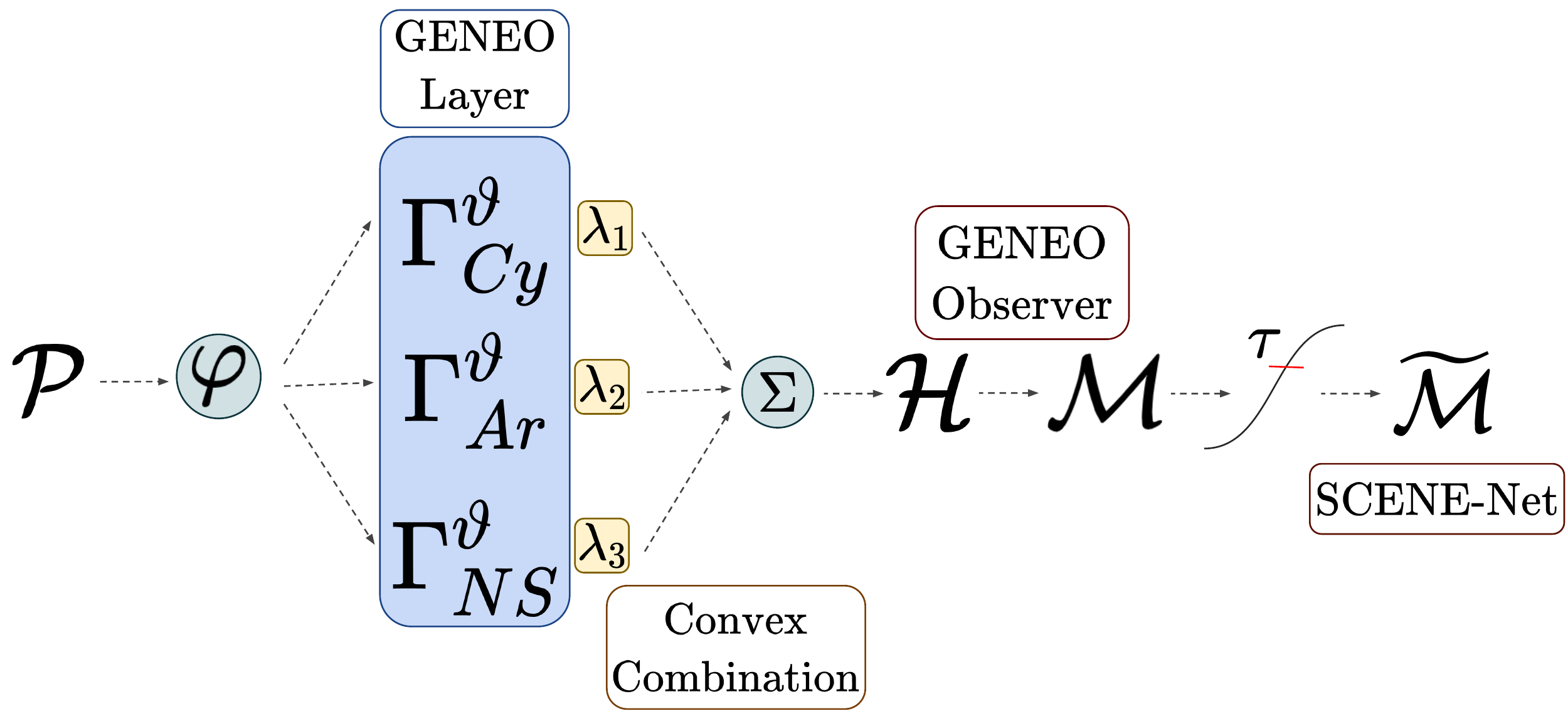}
    \caption{Pipeline of SCENE-Net: an input point cloud  $\mathcal{P}$ is measured according to function $\varphi$ and voxelized. This representation then is fed to a GENEO-layer, where each operator $\Gamma_i^{\vartheta_i}$ separately convolves the input.
    A GENEO observer $\mathcal{H}$ is then achieved by a convex combination of the operators in the GENEO layer.
    $\mathcal{M}$ transforms the analysis of the observer into a probability of belonging to a tower. %
    Lastly, a threshold operation is applied to classify the voxels. Note that this final step occurs after training is completed.
    }
    \label{fig:gnet_overview}
\end{figure}

\paragraph*{Overview}

3D Point clouds are generally denoted as $\mathcal{P} \in \R^{N\times (3 + d)}$, where  $N$ is the number of points and $3 + d$ is the cardinality of spatial coordinates plus any point-wise features, such as colors or normal vectors. 
%
%
The input point cloud is first transformed in accordance with a measurement function $\varphi\colon \R^3 \to \{0, 1\}$ that signals the presence of 3D points in a voxel discretization. 
Next, the transformed input is fed to a layer of multiple GENEOs (GENEO-layer), each chosen randomly from a parametric family of operators, and defined by a set of trainable shape parameters $\vartheta_i$ (Fig.~\ref{fig:gnet_overview}).
Such GENEOs are in the form of convolutional operators with carefully designed kernels as described later. Not only is convolution a well-studied operation, but it also offers equivariance w.r.t. translations by definition. 
During training, it is not the kernels themselves that are fine-tuned with back-propagation, since this would not preserve equivariance at each optimization step. Instead, the error is propagated to the shape parameters $\vartheta_i$ of each operator. 
Following the GENEO-layer, its set of operators $\Gamma=\{\Gamma_i^{\vartheta_i}\}_{i=1}^K$, 
with shape parameters $\vartheta = \vartheta_1,\dots,\vartheta_k$, 
are combined through convex combination with weights $\lambda = (\lambda_1,\dots,\lambda_k )^T$ by $\observer \colon \mathcal{P} \to \mathcal{P}$ such that
\begin{align}\label{eq:observer}
    \begin{split}
        &\observer(x) = \sum_{i=1}^{K}\lambda_i \Gamma_i^{\vartheta_i}(\varphi)(x)
        %
    \end{split}
\end{align}
Since the convex combination of GENEOs is also a GENEO~\cite{GENEO19}, $\mathcal{H}$ preserves the equivariance of each operator $\Gamma^\vartheta \in \Gamma$. 
In fact, $\mathcal{H}$ defines a GENEO observer that analyzes the 3D input scenes looking for the geometrical properties encoded in $\Gamma$.
The convex coefficients $\lambda$ represent the overall contribution of each operator $\Gamma_i^\vartheta$ to $\mathcal{H}$ to the analysis. 
The parameters grant our model its intrinsic interpretability. They are learned during training and represent geometric properties and the importance of each $\Gamma^\vartheta$ in modeling the ground truth.

Next, we transform the observer's analysis into a probability of each 3D voxel belonging to a supporting tower as a model $\model \colon \mathcal{P} \to [0, 1]^N$
\begin{align*}
    \begin{split}
        &\model(x) = \bigg(\tanh\Big(\observer(x)\Big)\bigg)_+
    \end{split},
\end{align*}
where $(t)_+ = \max\{0, t\}$ is the rectified linear unit (ReLU).
Negative signals in $\mathcal{H}(x)$ represent patterns that do not exhibit the sought-out geometrical properties. Conversely, positive values quantify their presence.
Therefore, $\tanh$ compresses the observer's value distribution into [-1, 1], and the ReLU is then applied to enforce a zero probability to negative signals.
%
Lastly, a probability threshold $\tau \in [0, 1]$ is defined through hyperparameter fine-tuning and applied to $\mathcal{M}$ resulting in a map $\widetilde{\model} \colon \mathcal{P}\times \R \to \{0, 1\}^N$
\begin{align*}
    \begin{split}
        &\widetilde{\model}(x, \tau) = \Big\{\model(x)\Big\} \geq \tau
    \end{split},
\end{align*}
where $\widetilde{\mathcal{M}}$ denotes the \textbf{SCENE-Net} model.

\paragraph*{Knowledge Engineering via GENEOs.}
In this section, we formally define the knowledge embedded in the observer $\mathcal{H}$. The following GENEOs describe power line supporting towers in order to fully discriminate them from their environment.

\paragraph*{Cylinder GENEO.}
The most striking characteristic of supporting towers against the rural environment is their long, vertical and narrow structure. 
As such, their identification is equivariant w.r.t. rotations along the \textit{z-axis} and translations in the \textit{xy} plane, which we encode by the means of a cylinder.

\begin{definition} \label{prop:cylinder}
In order to promote smooth patterns, a cylinder is defined by $g_{Cy}\colon \R^3 \to [0, 1]$:
\begin{align*}
    \begin{split}
        &g_{Cy}(x) = e^{-\frac{1}{2\sigma^2}(\Vert \pi_{-3}(x) - \pi_{-3}(c) \Vert^{2} - r^2 )^2}
    \end{split}
\end{align*}
%
where $\pi_{-3}(x) = (\pi_{1}(x), \pi_{2}(x),0)$ and $\pi_i$ defines a projection function of the $i$th element of the input vector. 
\end{definition}
Definition~\ref{prop:cylinder} is a smoothed characterization of the Cylinder defined in Appendix~\ref{supp_prop:cylinder}.
The function $g_{Cy}$ defines a smoothed cylinder centered in $c$ by means of a Gaussian function, with the distance between $x$ and the cylinder's radius ($r$) as its mean.
The shape parameters are the Gaussian's standard deviation and $r$, defined as $\vartheta_{Cy} = [r, \sigma]$.

\begin{figure}
    \centering
    \subfigure[Cylinder]{ \label{fig:cylinder}\includegraphics[width=.31\columnwidth]{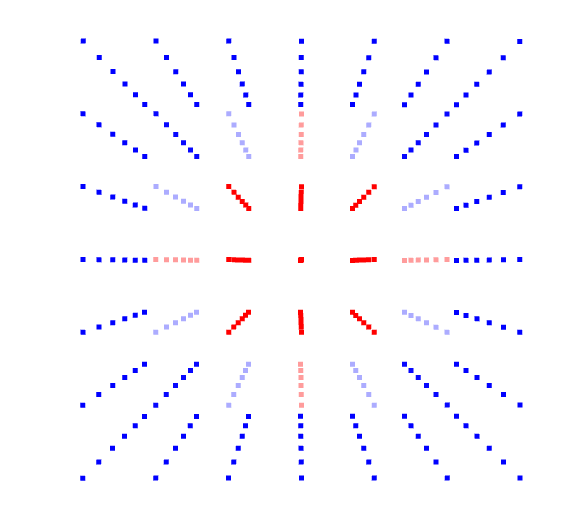}}
    \subfigure[Arrow]{ \label{fig:arrow}\includegraphics[width=.28\columnwidth]{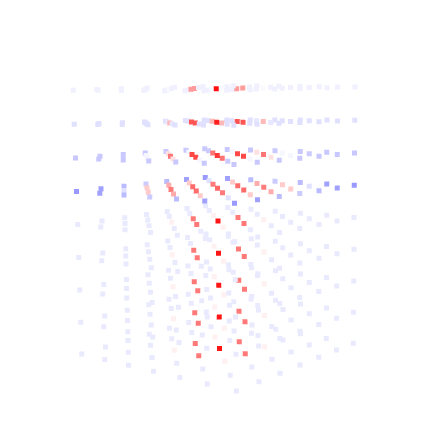}}
    \subfigure[Negative Sphere]{ \label{fig:neg_sphere}\includegraphics[width=.31\columnwidth]{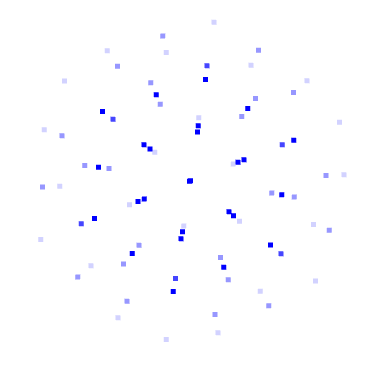}}
    \subfigure{\centering
    \includegraphics[width=.7\columnwidth]{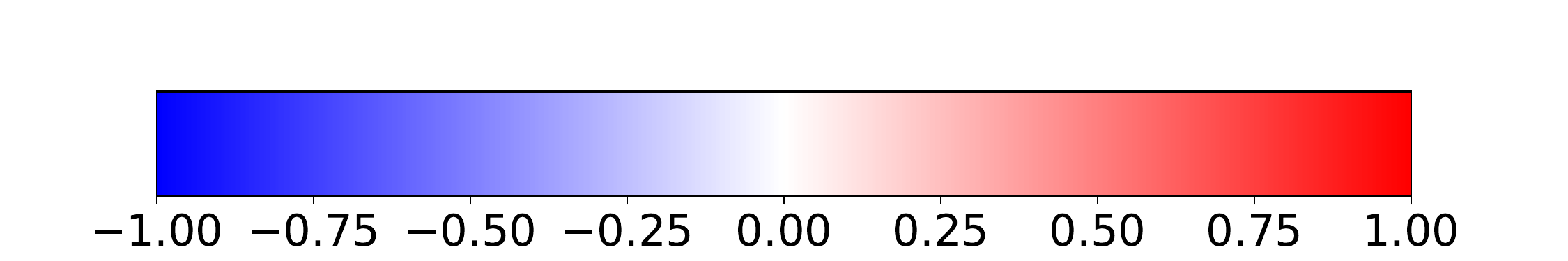}}
\caption{GENEO kernels discretized in a voxel grid and colored according to weight distribution.}
\end{figure}

GENEOs act on functions, transforming them to remain equivariant to a specific group of transformations. Our GENEOs act upon $\Phi$, the topological space representing $\mathcal{P}$ with admissible functions $\varphi\colon \R^3 \to \{0, 1\}$. 
Specifically, we work with appropriate $\varphi \in \Phi$ functions that represent point clouds and preserve their geometry. For instance, $\varphi$ can be a function that signals the presence of 3D points in a voxel grid.  
Therefore, the cylinder GENEO $\Gamma_{Cy}^\vartheta$ transforms $\varphi$ into a new function that detects sections in the input point cloud that demonstrate the properties of $g_{Cy}$ and, simultaneously, preserves the geometry of the 3D scene
\begin{align*}
        &\Gamma_{Cy}^\vartheta\colon \Phi \to \Psi, \qquad \psi_{Cy} = \Gamma_{Cy}^\vartheta (\varphi) \nonumber \\
        &\psi_{Cy} (x) = \int_{\R^3} \Tilde{g}_{Cy}(y)\varphi(x - y)dy
\end{align*}

where $\Psi$ is a new topological space that represents $\mathcal{P}$ with functions $\psi\colon \R^3 \to [0, 1]$ and $\Tilde{g}_{Cy}$ defines a normalized Cylinder.
The kernel $g_{Cy}$ is normalized to have a zero-sum to promote the stability of the observer.
This way, we encourage the geometrical properties that exhibit the sought-out group of transformations and punish those which do not.
Thus, $\psi_{Cy}(x)$ assumes positive values for 3D points near the radius, whereas negative values discourage shapes that do not fall under the $g_{Cy}$ definition. 
This leads to a more precise detection of the encoded group of transformations. 
The cylinder kernel discretized in a voxel grid can be seen in Fig.~\ref{fig:cylinder}.

\paragraph*{Arrow GENEO.}
Towers are not the only element in rural environments characterized by a vertical and narrow structure. The identification of trees also shows equivariance w.r.t. rotations along the \textit{z-axis}. 
Therefore, it is not enough to detect the body of towers, we also require the power lines that they support.
To this end, we define a cylinder following the rationale behind the cylinder GENEO with a cone on top of it.
This arrow defines equivariance w.r.t. the different angles at which power lines may find their supporting tower.
\begin{definition} \label{prop:arrow}
The function describing the Arrow is defined as $g_{Ar}\colon \R^{3} \to [0, 1]$:

\begin{align*}
    \begin{split}
        &g_{Ar}(x)=
        \twopartdef
        {e^{\frac{-1}{2\sigma^2}(\Vert \pi_{-3}(x) - \pi_{-3}(c) \Vert^{2} - r^2 )^2}}{\pi_3(x) < h}
        {e^{\frac{-1}{2\sigma^2}(\Vert \pi_{-3}(x) - \pi_{-3}(c) \Vert^{2} - (r_c\tan(\beta\pi))^2 )^2}}{\pi_3(x) \geq h}
    \end{split}
\end{align*}
with  $\beta \in [0, 0.5)$ defining the inclination of the cone.
\end{definition}
Definition~\ref{prop:arrow} is a smoothed characterization of the Arrow defined in Appendix~\ref{supp_prop:arrow}.
 The radii of the cylinder and cone are defined by $r$ and $r_c$, respectively, with $c$ as their center. Lastly, $h$ defines the height at which the cone is placed on top of the cylinder.
Thus, the shape parameters of the Arrow are defined by the vector $\vartheta_{Ar} = [r, \sigma, h, r_c, \beta].$
Lastly, we are also interested that this kernel sums to zero, so we define
\begin{align*}
        &\Gamma_{Ar}^\vartheta\colon \Phi \to \Psi, \qquad \psi_{Ar} = \Gamma_{Ar}^\vartheta (\varphi) \nonumber \\
        &\psi_{Ar} (x) = \int_{\R^3} \Tilde{g}_{Ar}(y)\varphi(x - y)dy,
\end{align*}
where $\Tilde{g}_{Ar}(y)$ represents a normalized Arrow kernel. Its discretization is depicted in Fig.~\ref{fig:arrow}.

\paragraph*{Negative Sphere GENEO.} 
Detecting power lines does not exclude the remaining objects in the scene whose identification also demonstrates equivariance w.r.t. rotations along the \textit{z-axis}. Tree elements, such as bushes, are especially frequent in the TS40K dataset. 
Thus, we designed a negative sphere to diminish their detection and simultaneously punish the geometry of trees.

\begin{definition} \label{prop:ns}
The Negative Sphere $g_{NS}\colon \R^3 \to [-\omega, 1[ $ is defined as
\begin{align*}
    \begin{split}
        &g_{NS}(x) = - \omega e^{\frac{-1}{2\sigma^2}(\Vert x - c\Vert^{2} - r^2 )^2}.
    \end{split}
\end{align*}
with $\omega \in ]0, 1]$ defining a small negative weight that punishes the spherical shape.
\end{definition}
Definition~\ref{prop:ns} is a smoothed characterization of the Negative Sphere in Appendix~\ref{supp_prop:ns}.
The shape parameters of this operator are $\vartheta_{NS} = [r, \sigma, \omega]$.
Since we wish to discourage spherical patterns following the definition of $g_{NS}$, so we do not enforce that its space sums to zero, obtaining
\begin{align*}
        &\Gamma_{NS}^\vartheta\colon \Phi \to \Psi_{NS}, \qquad \psi_{NS} = \Gamma_{NS}^\vartheta (\varphi) \nonumber \\
        &\psi_{NS} (x) = \int_{\R^3} g_{NS}(y)\varphi(x - y)dy.
\end{align*}
where $\Psi_{NS}$ is a topological space containing functions $\psi\colon \R^3 \to [-\omega, 1[$.
Fig.~\ref{fig:neg_sphere} depicts the computation of this kernel in a voxel grid.

\paragraph*{GENEO Loss.}

The use of GENEOs in knowledge embedding forces our model to uphold the convexity of the observer during training. 
Thus, our problem statement is represented by the following optimization problem
\begin{align*}
    \begin{split}
        \minimize_{\lambda, \vartheta}\quad &\expect_{X, y, \alpha, \epsilon}\bigg\{\mathcal{L}_{seg}(\lambda, \vartheta)\bigg\} \\
        \textrm{s.t.} \quad &\vartheta \geq 0 \\ \quad &\lambda^T\textbf{1} = 1\\ \quad &\lambda \geq 0,
    \end{split}
\end{align*}
where the segmentation loss $\mathcal{L}_{seg}$ is defined as
\begin{align*}
    &\mathcal{L}_{seg}(\lambda, \vartheta) =
    f_w(\alpha, \epsilon, y)\Big(\model(X) - y\Big)^2.
    %
\end{align*}
 The loss uses a weighted squared error following the weighting scheme $f_w$ proposed in~\cite{steininger2021density} to mitigate data imbalance. The hyperparameter $\alpha$ emphasizes the weighting scheme, whereas $\epsilon$ is a small positive number that ensures positive weights. Thus, $\expect\{\cdot \}$ represents the expectation of the segmentation loss over the data distribution.
The above constraints ensure that our model $\mathcal{M}$ maintains convexity throughout training, with \textbf{1} denoting a vector composed of entries one.
The reparametrization of the hyperparameters $\lambda$ to obtain an equivalent optimization problem, considering $\lambda_k = 1- \sum_{i=1}^{K-1} \lambda_i$, thus obtaining Problem~(\ref{eq:opt2}),
\begin{align}\label{eq:opt2}
    \begin{split}
        \minimize_{\lambda, \vartheta}\quad &\expect_{X, y, \alpha, \epsilon}\bigg\{{\mathcal{L}_{seg}}(\lambda, \vartheta)\bigg\} \\
        \textrm{s.t.} \quad &\vartheta \geq 0 \\ \quad &\lambda \geq 0
    \end{split} ,
\end{align}
allows for dropping one of the constraints.
Then, we ensure non-negativity of $\mathcal{M}$'s trainable parameters $\lambda, \vartheta$ by relaxing Problem~\eqref{eq:opt2} and introducing a penalty in the optimization cost definition as
\begin{align}\label{eq:opt_final}
    \begin{aligned}
        \minimize_{\lambda, \vartheta}&\quad \expect_{X, y, \alpha, \epsilon}\bigg\{{\mathcal{L}_{seg}}(\lambda, \vartheta)\bigg\}\; +
        \rho_l\Big( \sum_i^K h(\lambda_i)\Big) + \rho_t\Big( \sum_i^K \sum_j^{T_i} h(\vartheta_{ij})\Big),\\
    \end{aligned}
\end{align}
where $h(x) = \big( -x\big)_+ $, $\rho_l$ and $\rho_t$ are scaling factors of the negativity penalty $h$ and $T_i$ is the number of shape parameters in $\vartheta_i$.
GENEO final loss optimization is formalized in Problem~(\ref{eq:opt_final}). It consists of a data fidelity component (i.e., $\mathcal{L}_{seg}$) and two penalties on negative parameters.

\section{Experiments}
\label{sec:experiments}
%
%
In this Section, we assess properties of our model SCENE-Net that help electrical companies in the inspection of power lines: (1) interpretability of the model, (2) accuracy, (3) robustness to noisy labels, (4) training and inference time, and (5) performance on the SemanticKITTI benchmark.
%
Further details about the TS40K dataset, the training protocol, inference performance with high-resolution voxel grids, and ablation studies can be found in the Supplementary Material.

\begin{figure}
 \centering 
 \includegraphics[width=.9\columnwidth]{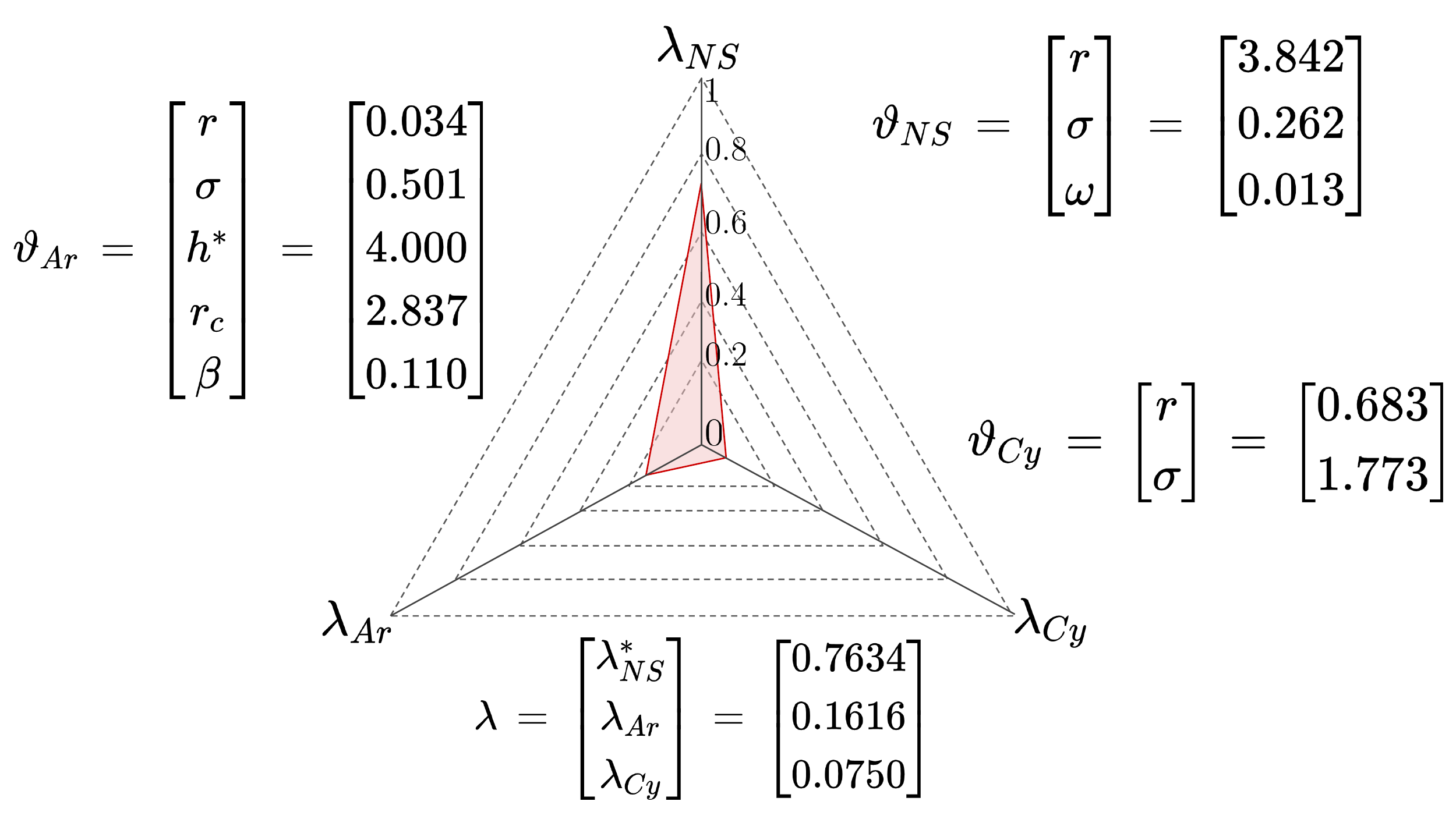}
 \caption{The trainable parameters of SCENE-Net, $\vartheta$ and $\lambda$. Parameter $h^*$ is not trainable, and $\lambda_{NS}^*$ is defined as a function of the other mixing weights $\lambda_{NS}^* = 1 - \lambda_{Ar} - \lambda_{Cy}$.}
 \label{fig:interpretable}
\end{figure}

\begin{figure}[h]
    \centering
     \subfigure[TS40K scene]{\includegraphics[width=.25\columnwidth]{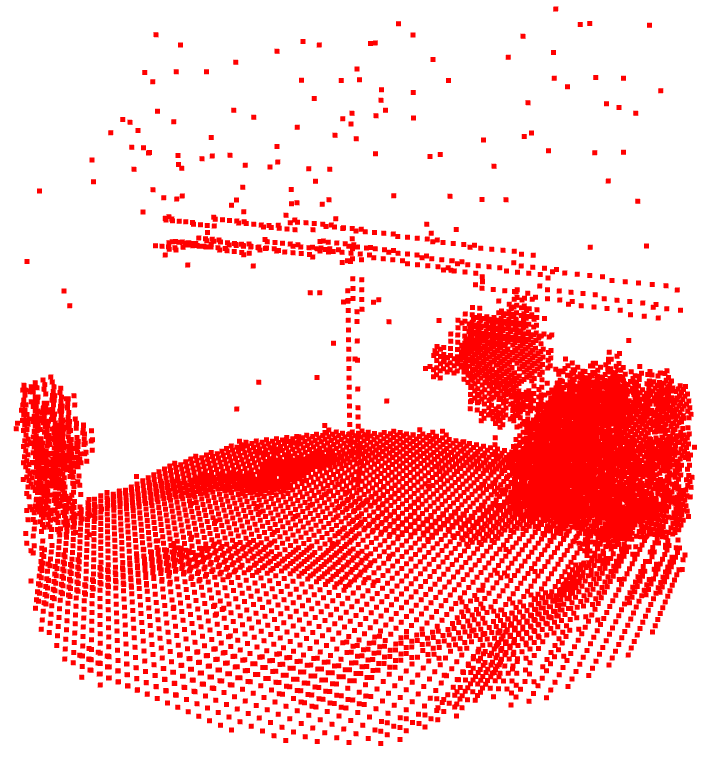}}
     \subfigure[Cylinder]
     {\includegraphics[width=.26\columnwidth]{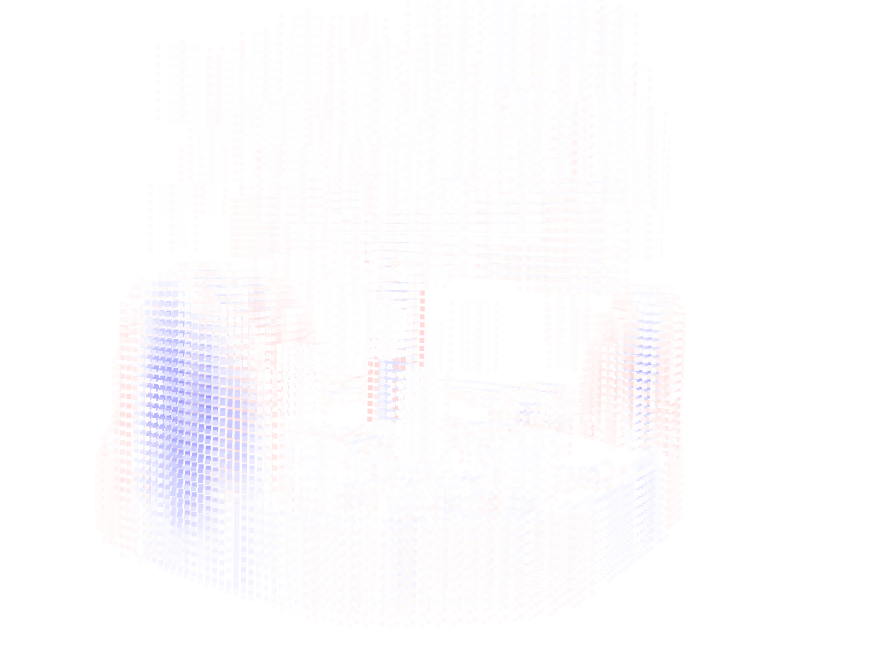}}
     \subfigure[Arrow]
     {\includegraphics[width=.25\columnwidth]{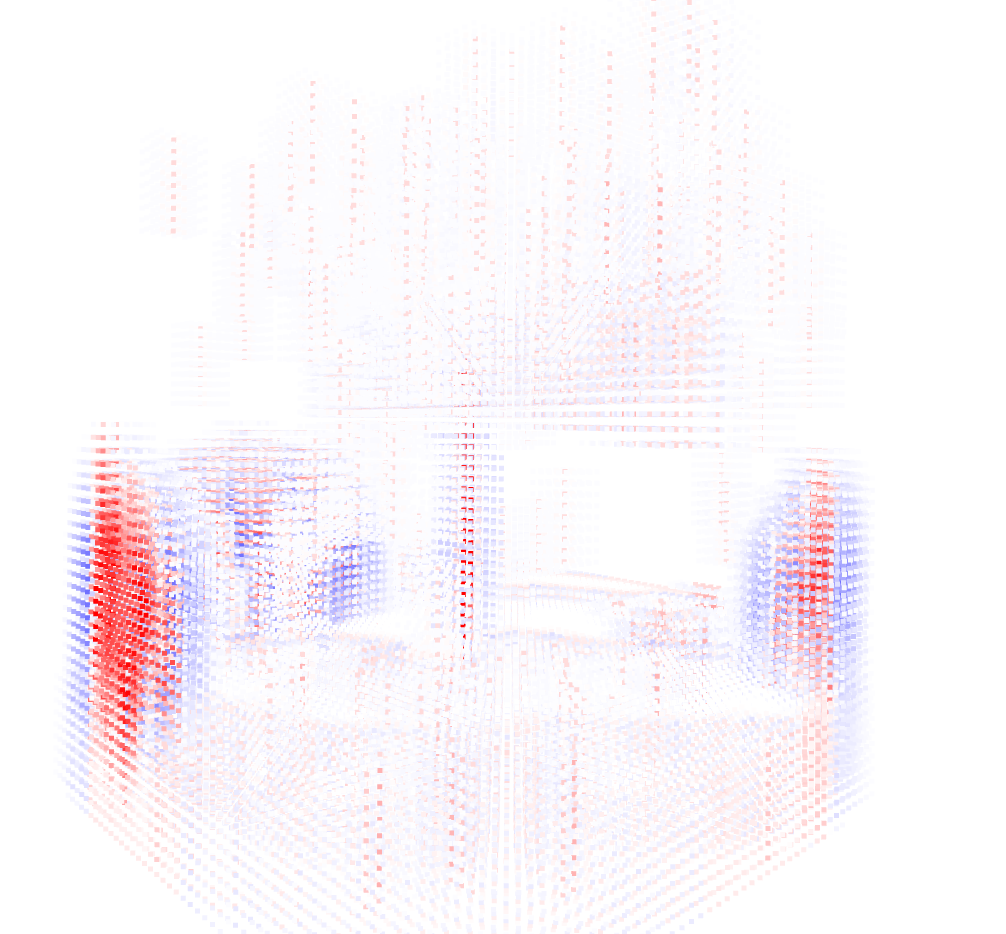}}
     \subfigure[Negative Sphere]
     {\includegraphics[width=.21\columnwidth]{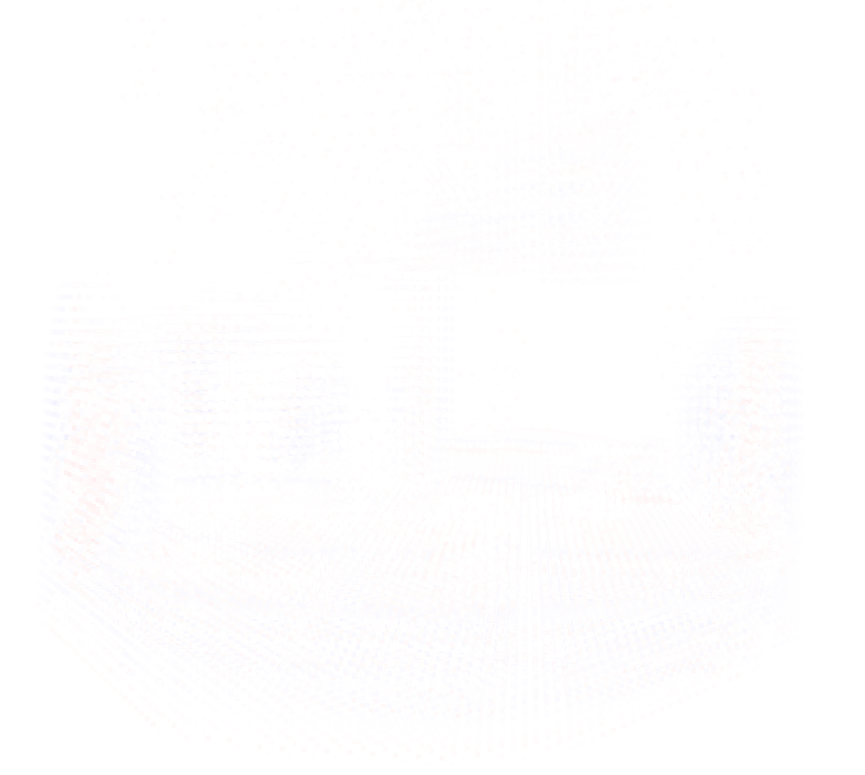}}
     \subfigure
     {\includegraphics[width=.7\columnwidth]{colorbar.pdf}}
     
    \caption{\textit{Post hoc} analysis of SCENE-Net. We can examine the activation of each geometric operator and correlate it to the detection of certain elements in the scene. We see that the Arrow is responsible for the most activation, while the Negative Sphere has a smaller absolute value.}
\label{fig:posthoc}
\end{figure}

\paragraph*{Interpretability of the trained SCENE-Net: The meaning of the 11 learned parameters.}
\label{sec:results-interpretability}
To understand if the model parameters are interpretable, we inspect SCENE-Net's 11 trainable parameters $\vartheta$ and $\lambda$ after training.
Each $\vartheta_i \in \vartheta$ holds the learned shape parameters of a geometrical operator $\Gamma_i$, such as their height or radius.
The convex coefficients $\lambda$ weigh each operator $\Gamma_i$ in our model's analysis.
For example, we can conclude that the instance $\vartheta_{NS}$ of the Negative Sphere GENEO ($\Gamma_{NS}$) holds a weight of 76.34\% on SCENE-Net's output (Fig.~\ref{fig:interpretable}).
The geometric nature of the observer and combination parameters endow intrinsic \textbf{interpretability} to SCENE-Net.  
\paragraph*{Post-hoc interpretation for specific predictions.}
%
We can correlate the detection of scene elements, such as vegetation, to the contributions of each GENEO. This provides an extra layer of transparency to our model.
%
The Arrow kernel is responsible for the detection of towers, the Cylinder aids this process and diminishes the detection of vegetation, and the Negative Sphere stabilizes the model by balancing contributions of the previous kernels (Fig~\ref{fig:posthoc}). 

\paragraph*{Qualitative accuracy and quantitative metrics: SCENE-Net is more precise in detecting towers than a baseline CNN.}
\label{sec:results-accuracy}
To evaluate if SCENE-Net can correctly identify towers in landscapes of the noisy TS40K dataset, we chose the task of 3D semantic segmentation of power line towers. We trained SCENE-Net and a baseline CNN according to the protocol described in the Supplementary Material. 
We use a CNN with similar architecture and the same base operator (i.e., convolution) for feature retrieval. 
The main difference is their kernel initialization: SCENE-Net kernels are randomly initialized, but belong to a precise family of operators, while CNN kernels are completely random.
%
Running models for 3D point cloud semantic segmentation~\cite{thomas2019kpconv,AF2S3Net,xu2021rpvnet,yan20222dpass} was not done due to their computational requirements.
The application penalizes the false positives more, thus we will emphasize Precision. Due to the imbalanced nature of the labels, we measured overall Precision, Recall, and Intersection over Union (IoU). Quantitatively, we observe a lift in Precision of 38\%, and of 5\% in IoU, and a drop of 13\% in Recall (Table~\ref{tab:res_GENEO}). 

\begin{table}[ht!]
\centering
\caption{3D semantic segmentation metrics on TS40K.} 
\begin{tabular}{llccc}
\hline
\multicolumn{2}{l}{Method}             & \multicolumn{1}{l}{Precision} & \multicolumn{1}{l}{Recall} & \multicolumn{1}{l}{IoU} \\ \hline
\multicolumn{2}{l}{CNN}                &  0.44 ($\pm$ 0.07)                      & \textbf{0.26} ($\pm$ 0.02)                     & 0.53                   \\
\multicolumn{2}{l}{SCENE-Net} & \textbf{0.82} ($\pm$ 0.08)                     & 0.13 ($\pm$ 0.05)                      & \textbf{0.58}                    \\ \hline
\end{tabular}
\label{tab:res_GENEO}
\end{table}

%
%
\begin{figure}[t]
 \centering 
 \includegraphics[width=.7\columnwidth]{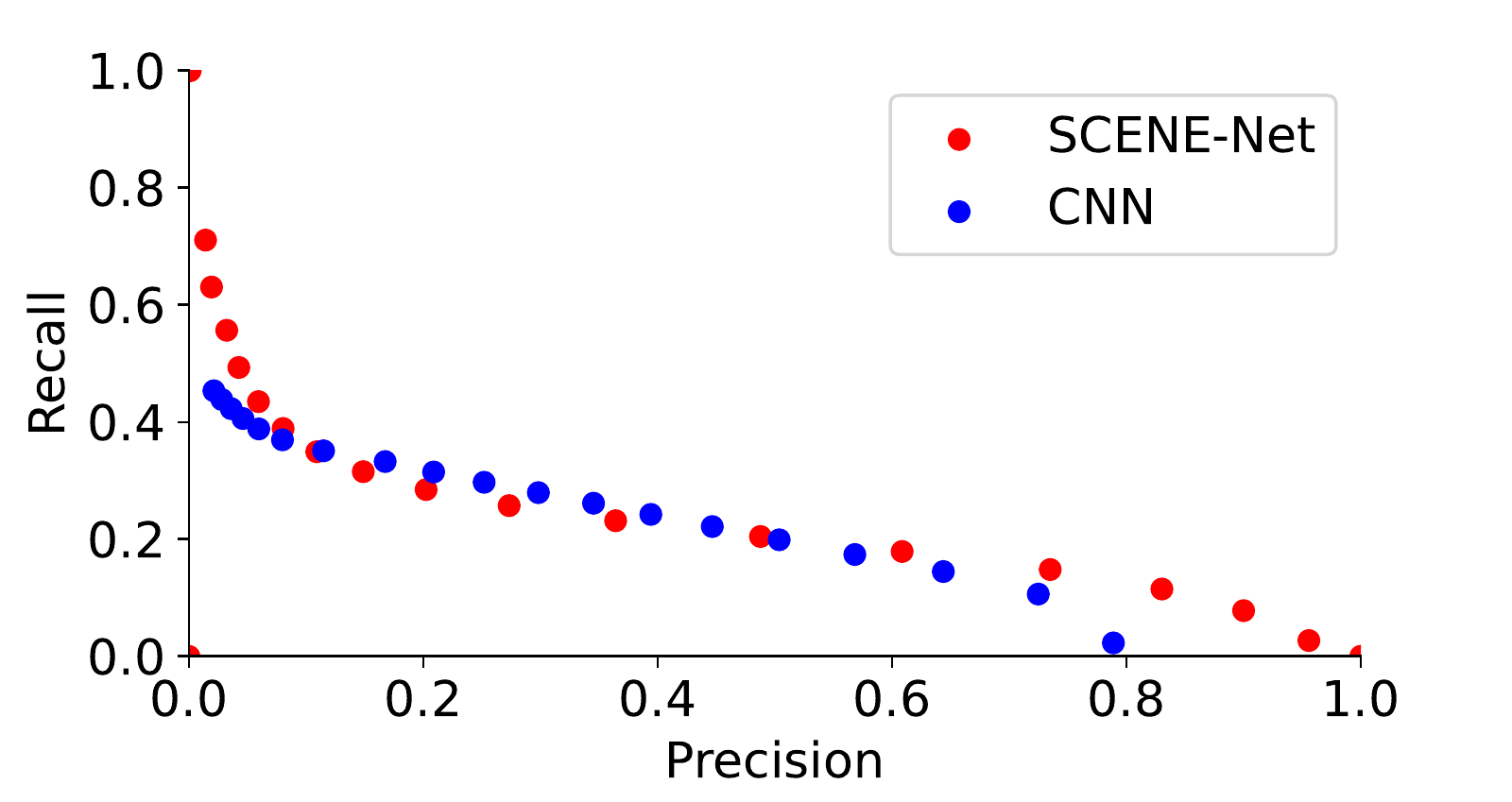}
 \caption{Precision-Recall curve for SCENE-Net and the CNN benchmark, with changing detection threshold. Although our model SCENE-Net has two orders of magnitude fewer parameters than the CNN, it attains a comparable area under the P-R curve.}
 \label{fig:recall}
\end{figure}

\begin{figure}[t]
\centering
    \subfigure[TS40K scene]
    {\label{fig:robust_input} \includegraphics[width=.5\textwidth]{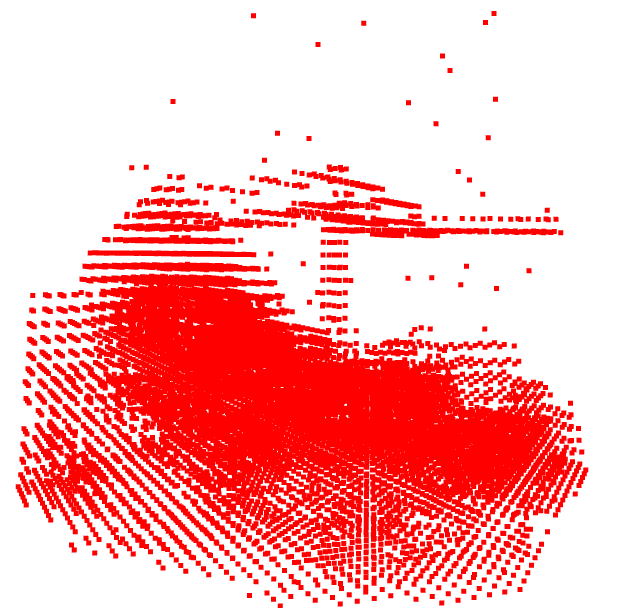}}
    \subfigure[SCENE-Net prediction against the GT]
     {\label{fig:robust_pred}
     \includegraphics[width=.25\textwidth]{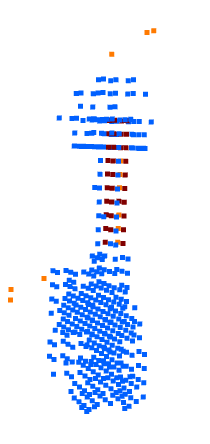}}
    \subfigure
     {\includegraphics[width=.5\columnwidth]
     {color_code_v2.pdf}}
    \caption{SCENE-Net is robust against mislabeled data. Fig.~\ref{fig:robust_pred} compares the prediction of SCENE-Net against the ground truth in Fig.~\ref{fig:robust_input}. SCENE-Net detects the body of the tower, ignoring the patch of ground mislabeled as a tower.}
\label{fig:robustness}
\end{figure}

The lower Recall of SCENE-Net is due to mislabeled points (Figs.~\ref{fig:intro_fig} and \ref{fig:robustness}),  and our choice to privilege Precision over Recall, in view of the fact that the Precision -- Recall curve is slightly better for SCENE-Net (Fig~\ref{fig:recall}).

\paragraph*{SCENE-Net is robust to noisy labels.}
\label{sec:results-robust}
It is important to assess the resilience to noisy labels in the Ground-Truth (GT) since 3D point clouds show more than 50\% of mislabeled points.
These examples are abundant in the dataset and SCENE-Net is able to recover the body of the tower without detecting ground and power line patches that are mislabeled as tower (Fig.~\ref{fig:robustness}). 
Most noisy labels on this kind of dataset are due to annotation excess around the object of interest and are not randomly distributed. These consistently incorrect labels entail low Recall values (Tab.~\ref{tab:res_GENEO}).

\paragraph*{SCENE-Net has low data requirements and has modest training time in common hardware.}
\label{sec:results-time-efficiency}

The design of SCENE-Net embedded with GENEO observers culminates in a model with 11 meaningful trainable parameters. 
This enables the use of common hardware (see Appendix~\ref{sec:train_protocal} for hardware specs) and a low data regime to train our model.  
The results reported in table~\ref{tab:sota_perf} were achieved with 5\% of the \textit{SemanticKITTI} training set. Training SCENE-Net with 50\% and 100\% of the available training data leads to a variation of 0.5\% in pole IoU performance.
Moreover, the number of parameters of SCENE-Net remains unchanged regardless of kernel size, whereas in traditional models, such as the baseline CNN with 2190 parameters, the number of parameters grows exponentially with larger kernel sizes.


\paragraph*{SCENE-Net on the SemanticKITTI: an efficient model for low-resource contexts.}

In Table~\ref{tab:sota_perf}, we present a comprehensive comparison of the performance of SCENE-Net against state-of-the-art models for the task of 3D semantic segmentation on the \textit{SemanticKITTI} benchmark, specifically in terms of pole IoU, number of parameters, and the ratio of pole IoU to number of parameters.
For this problem, we add to the GENEO loss in~(\ref{eq:opt_final}) the Tversky loss~\cite{salehi2017tversky} to boost IoU performance of SCENE-Net:
\begin{align*}
    \mathcal{L}_{Tversky}(y, \hat{y}) = 1 - \frac{y\hat{y} + \delta}{y\hat{y} + \alpha(\textbf{1} - y)\hat{y} + \beta y (\textbf{1} - \hat{y}) + \delta}
\end{align*}

where $y$, $\hat{y}$ are the ground truth and model prediction, $\alpha, \beta > 0$  are the penalty factors for false positives and false negatives respectively, and $\delta > 0$ is a smoothing term.

The comparison results demonstrate that SCENE-Net is a highly efficient model in terms of its parameter contribution. Our model has the lowest number of parameters, with at least a 5-order magnitude difference from the other models. 
SCENE-Net also has the highest ratio of pole IoU to a number of parameters, indicating that it can achieve a high level of performance with a minimal number of parameters. 
Although SCENE-Net does not achieve the highest pole IoU performance, it is somewhat on par with state-of-the-art models. 
Additionally, SCENE-Net provides intrinsic geometric interpretability and resource efficiency, which makes it a valuable model in high-risk tasks that require trustworthy predictions and good performance but have limited data and computing power.
%

\begin{table}[ht]
\centering
\caption{Semantic segmentation on \textit{SemanticKITTI}. Large models are included for comparison but cannot be used in low-resource contexts. Parameter efficiency is $\frac{\textrm{Pole IoU}}{\textrm{log \#Parameters}}$.}
\label{tab:sota_perf}
\begin{tabular}{cccc}
\hline
\multicolumn{1}{c}{Method} &
  \multicolumn{1}{c}{\begin{tabular}[c]{@{}c@{}}Pole\\ IoU\end{tabular}} &
  \multicolumn{1}{c}{\begin{tabular}[c]{@{}c@{}}\#Parameters\\ (M)\end{tabular}} &
  \multicolumn{1}{c}{\begin{tabular}[c]{@{}c@{}}Parameter \\ Efficiency\end{tabular}} \\ \hline
PointNet++~\cite{qi2017pointnet++}                & 16.9          & 1.48        & 1.19            \\
TangentConv~\cite{tatarchenko2018tangent}               & 35.8          & 0.4   & 2.77            \\
KPConv~\cite{thomas2019kpconv}                    & 56.4          & 14.9        & 3.41            \\
RandLA-Net~\cite{hu2020randla}                & 51.0          & 1.24            & 3.63            \\
RPVNet~\cite{xu2021rpvnet}                    & \textbf{64.8} & 24.8            & 3.80            \\
SparseConv~\cite{graham20183d}                    & 57.9          & 2.7         & 3.91            \\
JS3C-Net~\cite{yan2021sparse}                  & 60.7          & 2.7            & 4.09            \\
SPVNAS~\cite{tang2020searching}                    & 64.3          & 12.5       & 4.62            \\
\textbf{SCENE-Net (Ours)}                      & 57.5      & \textbf{1.1e-5}    & \textbf{23.98} \\ \hline
\end{tabular}%
\end{table}

\begin{figure}[t]
 \centering 
 \includegraphics[width=.7\textwidth]{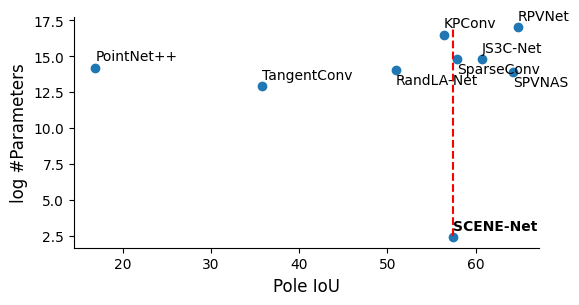}
 \caption{Semantic Segmentation results on the \textit{SemanticKITTI} benchmark. Log scale is used for more intelligible comparison.}
 \label{fig:semK_perf}
\end{figure}

\begin{figure}[t]
\centering
    \subfigure[\textit{SemanticKITTI} scene]
    {\label{fig:semK_input} \includegraphics[width=.45\columnwidth]{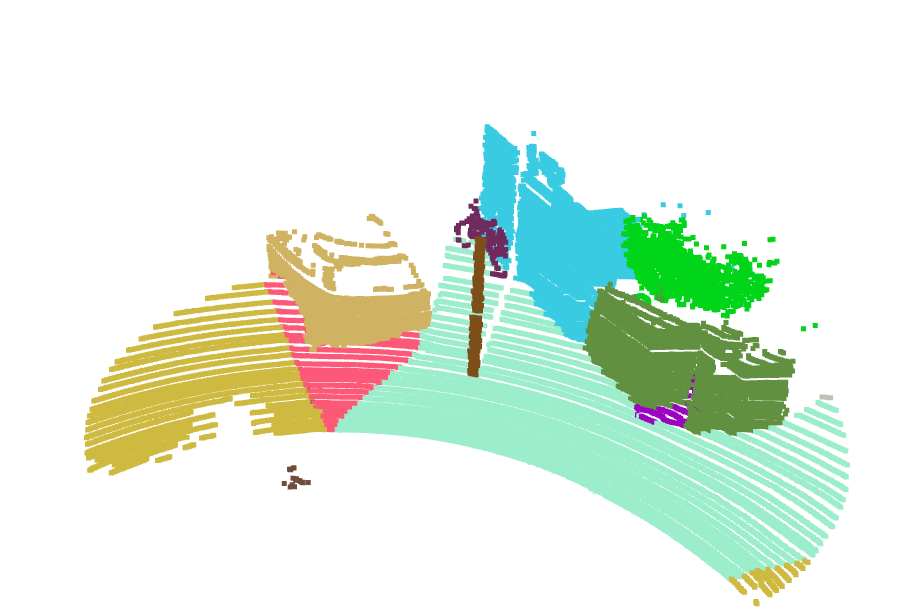}}
    \subfigure[SCENE-Net prediction against the GT]
     {\label{fig:semK_pred}
     \includegraphics[width=.5\columnwidth]{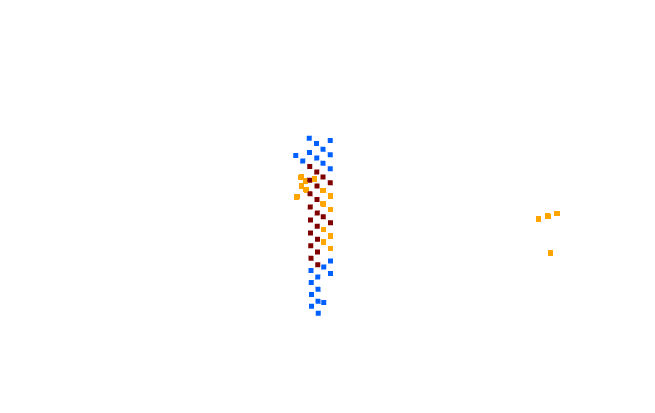}}
    \subfigure
     {\includegraphics[width=.5\columnwidth]
     {color_code_v2.pdf}}
    \caption{Qualitative results of SCENE-Net on \textit{SemanticKITTI} for pole detection. Fig.~\ref{fig:semK_pred} compares the prediction of SCENE-Net against the ground truth in Fig.~\ref{fig:semK_input}. SCENE-Net detects the body of the pole while disregarding the rest of the 3D scene.}
\label{fig:semK_qualitative}
\end{figure}

\section{Discussion}


Traditional companies, like utilities, need a resource-efficient, responsible application of ML models for the segmentation of real-world point clouds, e.g., for inspecting thousands of kilometers of a power grid.
In this paper, we present SCENE-Net, a low-resource white-box model for 3D semantic segmentation. Our approach offers a unique combination of intrinsic geometric interpretability, resource efficiency, and on-par state-of-the-art performance.
\paragraph*{Limitations}
Our model prioritizes transparency and performance over broader applicability while allowing a flexible extension.
In this paper, we segmented pole-like structures. State-of-the-art methods often show similar trade-offs, for example, 3D semantic segmentation models are tailored for autonomous driving~\cite{yan2021sparse,xu2021rpvnet}.
SCENE-Net requires a knowledge engineering phase that is not necessary for black-box models.
Despite these limitations, we believe that the transparency and efficiency of SCENE-Net make it a valuable tool for high-stakes applications.
To detect other shapes, other geometrical observers have to be created. As the convex combination of GENEOs is a GENEO, this problem is mitigated by creating a library of primary shapes to be combined to form more complex geometrical structures. This is relevant follow-up work, but out of the scope of this paper.
Multiclass and multilabel segmentation can be achieved by combining different binary class segmentation models.

\paragraph*{Impact}
%
%
From our experience deploying SCENE-Net within a utility company, low-resource transparent systems can critically help human decision-making---here, by facilitating fast and careful inspection of power lines with interpretable signals of observed geometrical properties. With only three observers and 11 meaningful trainable parameters, SCENE-Net can help reduce the risk of power outages and forest fires by learning from data.
%
%

%
%

%
%

%

\bibliography{biblio}

\begin{thebibliography}{10}

\bibitem{lipton2018mythos}
Zachary~C Lipton.
\newblock The mythos of model interpretability: In machine learning, the
  concept of interpretability is both important and slippery.
\newblock {\em Queue}, 16(3):31--57, 2018.

\bibitem{guidotti2018survey}
Riccardo Guidotti, Anna Monreale, Salvatore Ruggieri, Franco Turini, Fosca
  Giannotti, and Dino Pedreschi.
\newblock A survey of methods for explaining black box models.
\newblock {\em ACM computing surveys (CSUR)}, 51(5):1--42, 2018.

\bibitem{doshi2017towards}
Finale Doshi-Velez and Been Kim.
\newblock Towards a rigorous science of interpretable machine learning.
\newblock {\em arXiv preprint arXiv:1702.08608}, 2017.

\bibitem{rudin2019stop}
Cynthia Rudin.
\newblock Stop explaining black box machine learning models for high stakes
  decisions and use interpretable models instead.
\newblock {\em Nature Machine Intelligence}, 1(5):206--215, 2019.

\bibitem{Chefer_2021_CVPR}
Hila Chefer, Shir Gur, and Lior Wolf.
\newblock Transformer interpretability beyond attention visualization.
\newblock In {\em Proceedings of the IEEE/CVF Conference on Computer Vision and
  Pattern Recognition (CVPR)}, pages 782--791, June 2021.

\bibitem{voita2019analyzing}
Elena Voita, David Talbot, Fedor Moiseev, Rico Sennrich, and Ivan Titov.
\newblock Analyzing multi-head self-attention: Specialized heads do the heavy
  lifting, the rest can be pruned.
\newblock {\em arXiv preprint arXiv:1905.09418}, 2019.

\bibitem{ribeiro2016should}
Marco~Tulio Ribeiro, Sameer Singh, and Carlos Guestrin.
\newblock " why should i trust you?" explaining the predictions of any
  classifier.
\newblock In {\em Proceedings of the 22nd ACM SIGKDD international conference
  on knowledge discovery and data mining}, pages 1135--1144, 2016.

\bibitem{fong2017interpretable}
Ruth~C Fong and Andrea Vedaldi.
\newblock Interpretable explanations of black boxes by meaningful perturbation.
\newblock In {\em Proceedings of the IEEE international conference on computer
  vision}, pages 3429--3437, 2017.

\bibitem{chen2020concept}
Zhi Chen, Yijie Bei, and Cynthia Rudin.
\newblock Concept whitening for interpretable image recognition.
\newblock {\em Nature Machine Intelligence}, 2(12):772--782, 2020.

\bibitem{GENEO19}
Mattia~G Bergomi, Patrizio Frosini, Daniela Giorgi, and Nicola Quercioli.
\newblock Towards a topological--geometrical theory of group equivariant
  non-expansive operators for data analysis and machine learning.
\newblock {\em Nature Machine Intelligence}, 1(9):423--433, 2019.

\bibitem{GENEO21}
Pasquale Cascarano, Patrizio Frosini, Nicola Quercioli, and Amir Saki.
\newblock On the geometric and riemannian structure of the spaces of group
  equivariant non-expansive operators.
\newblock {\em arXiv preprint arXiv:2103.02543}, 2021.

\bibitem{Semantic3D}
Timo Hackel, Nikolay Savinov, Lubor Ladicky, Jan~D Wegner, Konrad Schindler,
  and Marc Pollefeys.
\newblock Semantic3d. net: A new large-scale point cloud classification
  benchmark.
\newblock {\em arXiv preprint arXiv:1704.03847}, 2017.

\bibitem{SemKITTI}
Jens Behley, Martin Garbade, Andres Milioto, Jan Quenzel, Sven Behnke, Cyrill
  Stachniss, and Jurgen Gall.
\newblock Semantickitti: A dataset for semantic scene understanding of lidar
  sequences.
\newblock In {\em Proceedings of the IEEE/CVF International Conference on
  Computer Vision}, pages 9297--9307, 2019.

\bibitem{muhammad2020deep}
Khan Muhammad, Amin Ullah, Jaime Lloret, Javier Del~Ser, and Victor Hugo~C
  de~Albuquerque.
\newblock Deep learning for safe autonomous driving: Current challenges and
  future directions.
\newblock {\em IEEE Transactions on Intelligent Transportation Systems},
  22(7):4316--4336, 2020.

\bibitem{alzubaidi2021review}
Laith Alzubaidi, Jinglan Zhang, Amjad~J Humaidi, Ayad Al-Dujaili, Ye~Duan,
  Omran Al-Shamma, Jos{\'e} Santamar{\'\i}a, Mohammed~A Fadhel, Muthana
  Al-Amidie, and Laith Farhan.
\newblock Review of deep learning: Concepts, cnn architectures, challenges,
  applications, future directions.
\newblock {\em Journal of big Data}, 8(1):1--74, 2021.

\bibitem{thomas2019kpconv}
Hugues Thomas, Charles~R Qi, Jean-Emmanuel Deschaud, Beatriz Marcotegui,
  Fran{\c{c}}ois Goulette, and Leonidas~J Guibas.
\newblock Kpconv: Flexible and deformable convolution for point clouds.
\newblock In {\em Proceedings of the IEEE/CVF International Conference on
  Computer Vision}, pages 6411--6420, 2019.

\bibitem{tang2020searching}
Haotian Tang, Zhijian Liu, Shengyu Zhao, Yujun Lin, Ji~Lin, Hanrui Wang, and
  Song Han.
\newblock Searching efficient 3d architectures with sparse point-voxel
  convolution.
\newblock In {\em European conference on computer vision}, pages 685--702.
  Springer, 2020.

\bibitem{xu2021rpvnet}
Jianyun Xu, Ruixiang Zhang, Jian Dou, Yushi Zhu, Jie Sun, and Shiliang Pu.
\newblock Rpvnet: A deep and efficient range-point-voxel fusion network for
  lidar point cloud segmentation.
\newblock In {\em Proceedings of the IEEE/CVF International Conference on
  Computer Vision}, pages 16024--16033, 2021.

\bibitem{yan2021sparse}
Xu~Yan, Jiantao Gao, Jie Li, Ruimao Zhang, Zhen Li, Rui Huang, and Shuguang
  Cui.
\newblock Sparse single sweep lidar point cloud segmentation via learning
  contextual shape priors from scene completion.
\newblock In {\em Proceedings of the AAAI Conference on Artificial
  Intelligence}, volume~35, pages 3101--3109, 2021.

\bibitem{yan20222dpass}
Xu~{Yan}, Jiantao {Gao}, Chaoda {Zheng}, Chao {Zheng}, Ruimao {Zhang}, Shenghui
  {Cui}, and Zhen {Li}.
\newblock {2DPASS: 2D Priors Assisted Semantic Segmentation on LiDAR Point
  Clouds}.
\newblock {\em arXiv e-prints}, page arXiv:2207.04397, July 2022.

\bibitem{long2015fully}
Jonathan Long, Evan Shelhamer, and Trevor Darrell.
\newblock Fully convolutional networks for semantic segmentation.
\newblock In {\em Proceedings of the IEEE conference on Computer Vision and
  Pattern Recognition}, pages 3431--3440, 2015.

\bibitem{rethage2018fully}
Dario Rethage, Johanna Wald, Jurgen Sturm, Nassir Navab, and Federico Tombari.
\newblock Fully-convolutional point networks for large-scale point clouds.
\newblock In {\em Proceedings of the European Conference on Computer Vision
  (ECCV)}, pages 596--611, 2018.

\bibitem{tchapmi2017segcloud}
Lyne Tchapmi, Christopher Choy, Iro Armeni, JunYoung Gwak, and Silvio Savarese.
\newblock Segcloud: Semantic segmentation of 3d point clouds.
\newblock In {\em 2017 International Conference on 3D vision (3DV)}, pages
  537--547. IEEE, 2017.

\bibitem{graham20183d}
Benjamin Graham, Martin Engelcke, and Laurens Van Der~Maaten.
\newblock 3d semantic segmentation with submanifold sparse convolutional
  networks.
\newblock In {\em Proceedings of the IEEE conference on Computer Vision and
  Pattern Recognition}, pages 9224--9232, 2018.

\bibitem{su2018splatnet}
Hang Su, Varun Jampani, Deqing Sun, Subhransu Maji, Evangelos Kalogerakis,
  Ming-Hsuan Yang, and Jan Kautz.
\newblock Splatnet: Sparse lattice networks for point cloud processing.
\newblock In {\em Proceedings of the IEEE conference on Computer Vision and
  Pattern Recognition}, pages 2530--2539, 2018.

\bibitem{wang2017cnn}
Peng-Shuai Wang, Yang Liu, Yu-Xiao Guo, Chun-Yu Sun, and Xin Tong.
\newblock {O-CNN}: Octree-based convolutional neural networks for 3d shape
  analysis.
\newblock {\em ACM Transactions On Graphics (TOG)}, 36(4):1--11, 2017.

\bibitem{le2018pointgrid}
Truc Le and Ye~Duan.
\newblock Pointgrid: A deep network for 3d shape understanding.
\newblock In {\em Proceedings of the IEEE conference on Computer Vision and
  Pattern Recognition}, pages 9204--9214, 2018.

\bibitem{qi2017pointnet}
Charles~R Qi, Hao Su, Kaichun Mo, and Leonidas~J Guibas.
\newblock Pointnet: Deep learning on point sets for 3d classification and
  segmentation.
\newblock In {\em Proceedings of the IEEE conference on Computer Vision and
  Pattern Recognition}, pages 652--660, 2017.

\bibitem{qi2017pointnet++}
Charles~Ruizhongtai Qi, Li~Yi, Hao Su, and Leonidas~J Guibas.
\newblock Pointnet++: Deep hierarchical feature learning on point sets in a
  metric space.
\newblock {\em Advances in neural information processing systems}, 30, 2017.

\bibitem{hu2020randla}
Qingyong Hu, Bo~Yang, Linhai Xie, Stefano Rosa, Yulan Guo, Zhihua Wang, Niki
  Trigoni, and Andrew Markham.
\newblock Randla-net: Efficient semantic segmentation of large-scale point
  clouds.
\newblock In {\em Proceedings of the IEEE/CVF Conference on Computer Vision and
  Pattern Recognition}, pages 11108--11117, 2020.

\bibitem{xu2020geometry}
Mingye Xu, Zhipeng Zhou, and Yu~Qiao.
\newblock Geometry sharing network for 3d point cloud classification and
  segmentation.
\newblock In {\em Proceedings of the AAAI Conference on Artificial
  Intelligence}, volume~34, pages 12500--12507, 2020.

\bibitem{hua2018pointwise}
Binh-Son Hua, Minh-Khoi Tran, and Sai-Kit Yeung.
\newblock Pointwise convolutional neural networks.
\newblock In {\em Proceedings of the IEEE conference on Computer Vision and
  Pattern Recognition}, pages 984--993, 2018.

\bibitem{li2018pointcnn}
Yangyan Li, Rui Bu, Mingchao Sun, Wei Wu, Xinhan Di, and Baoquan Chen.
\newblock {PointCNN}: Convolution on x-transformed points.
\newblock {\em Advances in Neural Information Processing Systems}, 31, 2018.

\bibitem{wu2019pointconv}
Wenxuan Wu, Zhongang Qi, and Li~Fuxin.
\newblock Pointconv: Deep convolutional networks on 3d point clouds.
\newblock In {\em Proceedings of the IEEE/CVF Conference on Computer Vision and
  Pattern Recognition}, pages 9621--9630, 2019.

\bibitem{Cylinder3D}
Xinge Zhu, Hui Zhou, Tai Wang, Fangzhou Hong, Yuexin Ma, Wei Li, Hongsheng Li,
  and Dahua Lin.
\newblock Cylindrical and asymmetrical 3d convolution networks for lidar
  segmentation.
\newblock In {\em Proceedings of the IEEE/CVF conference on Computer Vision and
  Pattern Recognition}, pages 9939--9948, 2021.

\bibitem{SensatUrban}
Qingyong Hu, Bo~Yang, Sheikh Khalid, Wen Xiao, Niki Trigoni, and Andrew
  Markham.
\newblock Towards semantic segmentation of urban-scale 3d point clouds: A
  dataset, benchmarks and challenges.
\newblock In {\em Proceedings of the IEEE/CVF conference on computer vision and
  pattern recognition}, pages 4977--4987, 2021.

\bibitem{AF2S3Net}
Ran Cheng, Ryan Razani, Ehsan Taghavi, Enxu Li, and Bingbing Liu.
\newblock 2-s3net: Attentive feature fusion with adaptive feature selection for
  sparse semantic segmentation network.
\newblock In {\em Proceedings of the IEEE/CVF conference on Computer Vision and
  Pattern Recognition}, pages 12547--12556, 2021.

\bibitem{guo2020deep}
Yulan Guo, Hanyun Wang, Qingyong Hu, Hao Liu, Li~Liu, and Mohammed Bennamoun.
\newblock Deep learning for 3d point clouds: A survey.
\newblock {\em IEEE transactions on pattern analysis and machine intelligence},
  43(12):4338--4364, 2020.

\bibitem{ribeiro2018anchors}
Marco~Tulio Ribeiro, Sameer Singh, and Carlos Guestrin.
\newblock Anchors: High-precision model-agnostic explanations.
\newblock In {\em Proceedings of the AAAI conference on artificial
  intelligence}, volume~32, 2018.

\bibitem{leite21}
Manuel Ribeiro and Joao Leite.
\newblock Aligning artificial neural networks and ontologies towards
  explainable ai.
\newblock In {\em Proceedings of the AAAI Conference on Artificial
  Intelligence}, volume~35, pages 4932--4940, 2021.

\bibitem{Zhang_2018_CVPR}
Quanshi Zhang, Ying~Nian Wu, and Song-Chun Zhu.
\newblock Interpretable convolutional neural networks.
\newblock In {\em Proceedings of the IEEE Conference on Computer Vision and
  Pattern Recognition (CVPR)}, June 2018.

\bibitem{bocchi2022geneonet}
Giovanni Bocchi, Patrizio Frosini, Alessandra Micheletti, Alessandro Pedretti,
  Carmen Gratteri, Filippo Lunghini, Andrea~Rosario Beccari, and Carmine
  Talarico.
\newblock Geneonet: A new machine learning paradigm based on group equivariant
  non-expansive operators. an application to protein pocket detection.
\newblock {\em arXiv preprint arXiv:2202.00451}, 2022.

\bibitem{conti2022construction}
Francesco Conti, Patrizio Frosini, and Nicola Quercioli.
\newblock On the construction of group equivariant non-expansive operators via
  permutants and symmetric functions.
\newblock {\em Frontiers in Artificial Intelligence}, 5:16, 2022.

\bibitem{botteghi2020finite}
Stefano Botteghi, Martina Brasini, Patrizio Frosini, and Nicola Quercioli.
\newblock On the finite representation of group equivariant operators via
  permutant measures.
\newblock {\em arXiv preprint arXiv:2008.06340}, 2020.

\bibitem{steininger2021density}
Michael Steininger, Konstantin Kobs, Padraig Davidson, Anna Krause, and Andreas
  Hotho.
\newblock Density-based weighting for imbalanced regression.
\newblock {\em Machine Learning}, 110(8):2187--2211, 2021.

\bibitem{salehi2017tversky}
Seyed Sadegh~Mohseni Salehi, Deniz Erdogmus, and Ali Gholipour.
\newblock Tversky loss function for image segmentation using 3d fully
  convolutional deep networks.
\newblock In {\em Machine Learning in Medical Imaging: 8th International
  Workshop, MLMI 2017, Held in Conjunction with MICCAI 2017, Quebec City, QC,
  Canada, September 10, 2017, Proceedings 8}, pages 379--387. Springer, 2017.

\bibitem{tatarchenko2018tangent}
Maxim Tatarchenko, Jaesik Park, Vladlen Koltun, and Qian-Yi Zhou.
\newblock Tangent convolutions for dense prediction in 3d.
\newblock In {\em Proceedings of the IEEE Conference on Computer Vision and
  Pattern Recognition}, pages 3887--3896, 2018.

\bibitem{ding2021electric}
Leiqing Ding, Jianjun Wang, and Yikai Wu.
\newblock Electric power line patrol operation based on vision and laser slam
  fusion perception.
\newblock In {\em 2021 IEEE 4th International Conference on Automation,
  Electronics and Electrical Engineering (AUTEEE)}, pages 125--129. IEEE, 2021.

\bibitem{guo2019research}
Tao Guo, Lianggang Xu, Yuran Chen, Yong Liu, Jian Zhao, Xiao Luo, and Shaohua
  Wu.
\newblock Research on point cloud power line segmentation and fitting
  algorithm.
\newblock In {\em 2019 IEEE 4th Advanced Information Technology, Electronic and
  Automation Control Conference (IAEAC)}, volume~1, pages 2404--2409. IEEE,
  2019.

\bibitem{tao2019study}
Guo Tao, Xu~Lianggang, Yang Heng, Yang Liugui, Zhao Jian, Wu~Shaohua, and Wang
  Di.
\newblock Study on segmentation algorithm with missing point cloud in power
  line.
\newblock In {\em 2019 IEEE 3rd Advanced Information Management, Communicates,
  Electronic and Automation Control Conference (IMCEC)}, pages 1895--1899.
  IEEE, 2019.

\end{thebibliography}
\bibliographystyle{unsrt}


\clearpage
\newpage
\appendix

\section{Signature Shape Definitions}
Although functions $g$ are straightforward to understand and deduce, for completeness we present the exact signature shape definitions from which functions $g$ were derived.

\subsection{Cylinder Definition}

\begin{definition}\label{supp_prop:cylinder}
The most striking characteristic of supporting towers against the rural environment is their long, vertical and narrow structure. 
As such, their detection is equivariant w.r.t. both rotations along the \textit{z-axis} and translations in the \textit{xy} plane, which we encode by means of a cylinder:
%
\begin{align*}
        &f_{Cy}\colon \R^{3} \to \{0, 1\}\\
        &f_{Cy}(x)= \twopartdef{1}{\Vert \pi_{-3}(x) -\pi_{-3}(c) \Vert^{2} = r^2}
        {0}{ \text{otherwise}},
\end{align*}

where $\pi_i$ defines a projection function of the $i$th element of the input vector. However, data patterns show smoothing, so we relax this condition by defining $g_{Cy}\colon \R^3 \to [0, 1]$:
\begin{align}
    \begin{split}
        &g_{Cy}(x) = e^{-\frac{1}{2\sigma^2}(\Vert \pi_{-3}(x) - \pi_{-3}(c) \Vert^{2} - r^2 )^2}
    \end{split}.
\end{align}
The function $g_{Cy}$ defines a cylinder centered in $c$ by means of a Gaussian function, with the distance between $x$ and the cylinder's radius ($r$) as its mean.
The shape parameters of the Cylinder are the standard deviation of the Gaussian and radius $r$ and are defined by $\vartheta_{Cy} = [r, \sigma]$.
\end{definition}

\subsection{Arrow Definition}

\begin{definition} \label{supp_prop:arrow}

Towers are not the only element in rural environments characterized by a vertical and narrow structure. For example, the detection of trees is also equivariant w.r.t. rotations along the \textit{z-axis}. 
Therefore, it is not enough to detect the body of towers, we also require the power lines that they support.
To this end, we define a cylinder following the rationale behind the Cylinder GENEO with a cone on top of it.
This arrow defines equivariance w.r.t. the different angles that power lines may have with a supporting tower. Formally, this function can be defined as $f_{Ar}\colon \R^{3} \to \{0, 1\}$
\begin{align*}
    \begin{split}
        &f_{Ar}(x)=
        \threepartdef{1}{\Vert \pi_{-3}(x) - \pi_{-3}(c) \Vert^{2} = r^2\\ &\wedge \pi_3(x) < h}
        {1}{\Vert \pi_{-3}(x) - \pi_{-3}(c) \Vert^{2} = (r_c\tan(\beta\pi))^2\\ &\wedge \pi_3(x) \geq h}
        {0}{\text{otherwise}},
    \end{split}
\end{align*}
with $\beta \in [0, 0.5)$ defining the cone's inclination. The cylinder and cone radii are defined by $r$ and $r_c$, respectively, with $c$ as their center. Lastly, $h$ defines the height at which the arrow is placed on top of the cylinder. This definition is too strict to yield any feasible results in real-world scenarios, so we smooth the conditions as follows
\begin{align}
    \begin{split}
        &g_{Ar}\colon \R^{3} \to [0, 1]\\
        &g_{Ar}(x)=
        \twopartdef
        {e^{\frac{-1}{2\sigma^2}(\Vert \pi_{-3}(x) - \pi_{-3}(c) \Vert^{2} - r^2 )^2}}{\pi_3(x) < h}
        {e^{\frac{-1}{2\sigma^2}(\Vert \pi_{-3}(x) - \pi_{-3}(c) \Vert^{2} - (r_c\tan(\beta\pi))^2 )^2}}{\pi_3(x) \geq h}
    \end{split}.
\end{align}
Thus, the shape parameters of the Arrow are defined by the vector $\vartheta_{Ar} = [r, \sigma, h, r_c, \beta].$
\end{definition}

\subsection{Negative Sphere Definition}

\begin{definition} \label{supp_prop:ns}

Detecting power lines does not exclude the remaining objects in the scene whose detection also demonstrates equivariance w.r.t. rotations along the \textit{z-axis}. Arboreal elements are especially frequent in the TS40K dataset. 
Thus, we designed a negative sphere to diminish their detection and, simultaneously, punish the geometry of trees $f_{NS}\colon \R^{3} \to \{-\omega, 1\}$ as
\begin{align*}
    \begin{split}
        &f_{NS}(x)=
        \twopartdef
        {1}{\Vert x - c\Vert^{2} = r^2}
        {-\omega}{\text{otherwise}},
    \end{split}
\end{align*}
with $\omega \in ]0, 1]$ defining a small negative weight that punishes the spherical shape. Next, we proceed with the relaxation of the sphere definition with function $g_{NS}$ as
\begin{align}
    \begin{split}
        &g_{NS}\colon \R^3 \to [-\omega, 1[ \\
        &g_{NS}(x) = e^{\frac{-1}{2\sigma^2}(\Vert x - c\Vert^{2} - r^2 )^2} - \omega.
    \end{split}
\end{align}
The shape parameters of this operator are defined as $\vartheta_{NS} = [r, \sigma, \omega]$. 
\end{definition}

\section{TS40K Dataset \label{sec:ts40k}}

\begin{figure*}[t]
    \centering
    \includegraphics[width=1\textwidth]{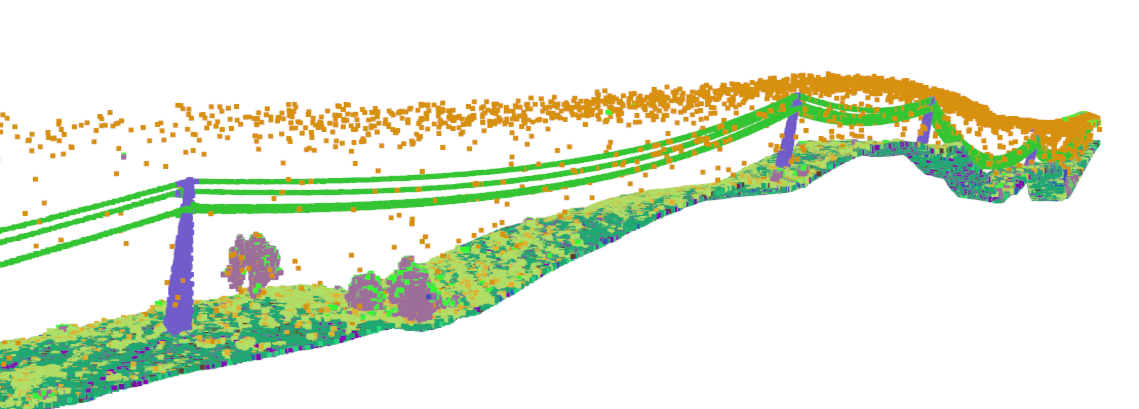}
    \caption{Visualization of TS40K raw point cloud with colored labels.}
    \label{fig:ts40k}
\end{figure*}

Electrical companies are responsible for the maintenance and inspection of the transmission system. They deploy low-flying helicopters to scan rural environments, from a BEV perspective, where the electrical grid is located.
The produced point clouds exhibit different data properties when compared to 3D scenes captured from other viewpoints, such as from a vehicle.
Namely, they show high point density and no object occlusion, scene elements present homogeneous density and no sparsity.
Then, the acquired 3D data is processed by maintenance personnel. Specifically, as the raw point clouds are dense and mainly encompass rural areas, data is sectioned into strips of land focused on the transmission system (Figure~\ref{fig:ts40k}). 
The raw data comprises several LiDAR files containing roughly 40 000 kilometers of the above land strips. 3D points therein are labeled with one out of 22 possible classes, such as power lines and their supporting towers, low and medium vegetation, rivers, railroads, human-made structures that do not belong to the transmission network, the ground, and optic cables, among others.
Table~\ref{tab:ts40k_labels} depicts these classes and their density in the dataset. Rail lines and road surfaces constitute the majority of the dataset (63\%), whereas classes of interest, such as power line supporting towers, make up less than 5\% of the overall data. 

\begin{table}[]
\caption{Available classes in the TS40K dataset and their distribution. Rail lines and road surfaces constitute the majority of the dataset (63\%). Our class of interest, power line towers, only makes up 0.52\%. 
Moreover, around 40\% of the tower points are mislabelled.\\}
\resizebox{\columnwidth}{!}{%
\begin{tabular}{lll|lll}
\hline
\textbf{Label} &
  \textbf{Class} &
  \textbf{Density(\%)} &
  \textbf{Label} &
  \textbf{Class} &
  \textbf{Density(\%)} \\ \hline
0  & Created          & 0               & 11 & Road surface                    & \textbf{18.758} \\
1  & Unclassified     & 0.571           & 12 & Overlap points                  & 23.403          \\
2  & Ground           & 0.529           & 13 & Medium Reliability              & 0               \\
3  & Low vegetation   & 0.681           & 14 & Low Reliability                 & 0               \\
4 &
  Medium vegetation &
  0.241 &
  {\color[HTML]{CB0000} 15} &
  {\color[HTML]{CB0000} Power line support tower} &
  {\color[HTML]{CB0000} \textbf{0.519}} \\
5  & Natural obstacle & 1.069           & 16 & Main power line                 & 0.907           \\
6  & Human structures & 0               & 17 & Other power line                & 0.002           \\
7  & Low point        & 0.362           & 18 & Fiber optic cable               & 0               \\
8  & Model key points  & 0               & 19 & Not rated object to be consider & 8.205           \\
9  & Water            & 0               & 20 & Not rated object to be ignored  & 0               \\
10 & Rail             & \textbf{44.752} & 21 & Incidents                       & 0               \\ \hline
\end{tabular}%
}
\label{tab:ts40k_labels}
\end{table}

\section{Related Work}

\subsection{Power line segmentation from 3D point clouds.} 
Power line inspection is generally performed by on-site maintenance personnel and manned helicopters that examine the power grid with portable devices or the naked eye. 
These methods are costly, inefficient, and demanding for staff. 
Thus, process automation is crucial for operators.
To this end, unmanned aerial vehicles (UAVs) carrying LiDAR sensors are deployed to scan the power grid and capture a 3D point cloud representation of the environment.
In the work~\cite{ding2021electric}, the authors combine SLAM algorithms with multi-sensor data to patrol the electrical grid with UAVs. This method employs a multi-view-based approach to point cloud segmentation, so the 3D reconstructions from 2D raster maps usually introduce information loss and decrease in accuracy.
Alternative methods project point clouds onto the $xy$-plane in order to cluster them~\cite{guo2019research} to segment power lines. This approach does not consider ground and irregular terrain and focuses on segmenting incomplete power lines.
Other methods take advantage of fine-grained elevation statistics of the original point cloud and $xy$-plane projections~\cite{tao2019study}.
The proposals above focus on the segmentation of high-voltage power lines and disregard their supporting towers.
By incorporating prior knowledge, our proposal segments supporting towers of any voltage. Not only are these structures also subject to inspections, but they serve as a point of reference for the location of power lines. 
In addition, by taking into account raw 3D scenes, other scene elements, such as vegetation, can be segmented to assess the risk of contact between the power grid and the environment.

%
%

\section{Experiments}

\subsection{Dataset.}


We evaluate the effectiveness of SCENE-Net on the TS40K dataset.
To mitigate the severe data imbalance in our class of interest, we created an ancillary dataset focused on power line-supporting towers with 2823 samples. For each tower in the 3D data, we crop the ground around it with a radius equal to its height.
This process introduces bias on classical machine learning agents, such as CNNs.
Contrastingly, SCENE-Net is optimized, after incorporating the appropriate prior knowledge, to detect the chosen features that describe the GT. 
A biased training dataset results in an agent tailored to detect supporting towers' geometry. Elements in the 3D scene that do not align with these properties are not signaled by SCENE-Net.
Furthermore, we discretize the cropped point clouds with a volume of $64^3$ voxels. If a voxel contains any 3D tower points, it is labeled as a tower voxel, otherwise, it is labeled as a non-tower voxel. This emphasizes the geometry of supporting towers so that they can be better described by SCENE-Net. 
To represent the 3D input, all non-empty voxels are given a value of 1. This measurement function preserves the structure of raw point clouds and mitigates the difference in point density between supporting towers and other classes. 

\subsection{Training Protocol.\label{sec:train_protocal}}

During the end-to-end training process of SCENE-Net, we adopt the following settings:
batch size is 8 for a total of 50 epochs. We employ the RMSProp optimizer with a learning rate of 0.001. The weighting scheme parameters $\alpha$ and $\epsilon$ are set to 5 and 0.1, respectively. While both scaling factors of the non-positive penalty $\rho_l$ and $\rho_t$ are set to 5. The kernel size used to discretize the GENEO operators is $9^3$.
The GENEOs parameters $\vartheta$ are randomly initialized under suitable and positive ranges. While the convex coefficients $\lambda_0,\dots,\lambda_{n-1}$ are randomly initialized in the range $[0, \frac{2}{N-1}]$ to promote a valid convex space for $\mathcal{H}$.
To demonstrate that SCENE-Net achieves good results even with fewer data, we use 20\% of the dataset for training, 10\% for validation, and 70\% for testing.
All experiments were conducted on an NVIDIA GeForce RTX 3070 GPU.

\subsection{Ablation Studies.}

\begin{table}[h]
\centering
\caption{Ablation Study of SCENE-Net on TS40K validation set.}
\begin{tabular}{cl|ccc|ccc}
\hline
\multicolumn{2}{c|}{Model} & Cylinder & Arrow & Neg. Sphere & Precision & Recall & IoU  \\ \hline
\multicolumn{2}{c|}{A}     & 1        & 0    & 0           & 0         & 0      & 0 \\
\multicolumn{2}{c|}{B}     & 0        & 1    & 0           & 0         & 0      & 0 \\
\multicolumn{2}{c|}{C}     & 1        & 0    & 1           & 0.34      & 0.01   & 0.12 \\
\multicolumn{2}{c|}{D}     & 0        & 1    & 1           & 0.13      & 0.01   & 0.08 \\
\multicolumn{2}{c|}{{\color[HTML]{000000} \textbf{E (Ours)}}} &
  {\color[HTML]{000000} \textbf{1}} &
  {\color[HTML]{000000} \textbf{1}} &
  {\color[HTML]{000000} \textbf{1}} &
  {\color[HTML]{000000} \textbf{0.82}} &
  {\color[HTML]{000000} \textbf{0.13}} &
  {\color[HTML]{000000} \textbf{0.58}} \\
\multicolumn{2}{c|}{F}     & 2        & 2    & 2           & 0.56      & 0.16   & 0.53 \\
\multicolumn{2}{c|}{G}     & 3        & 3    & 3           & 0.37      & 0.22   & 0.56 \\ \hline
\end{tabular}%
\label{tab:ablation}
\end{table}

In this section, we conduct ablation studies on SCENE-Net's architecture, specifically on the number of instances of each GENEO. All ablated models were tested on the TS40K validation set.
Table~\ref{tab:ablation} shows the following results:
Models A and B are each equipped with a single GENEO, demonstrating an overall poor performance. The Negative Sphere (NS) GENEO proved essential for our observer to disregard arboreal elements in the scene. 
Models C and D study if employing the Cylinder or Arrow combined with NS is enough to analyze the TS40K scenes.
However, SCENE-Net's architecture (model E) yields better results.
Lastly, models F and G test the use of multiple instances of each GENEO, however, this proved to decrease performance when compared to model E.

\subsection{Post-hoc interpretation of SCENE-Net  analysis.}

The geometric operators that embed SCENE-Net with prior knowledge not only provide intrinsic interpretability through its parameters $\vartheta$ and $\lambda$ but also enable a post-hoc interpretation of SCENE-Net's predictions. 
Specifically, we can analyze the observations of each GENEO operator separately before their convex combination. 
This way, we can correlate the detection or lack thereof of certain scene elements, such as bushes and towers, to the contributions of each GENEO operator. This provides an extra layer of transparency to our model and facilitates the knowledge engineering phase of GENEO-powered models.
%


\subsection{SCENE-Net inference in high resolution, when trained with low-resolution kernel sizes.}
\label{sec:results-resolution}
\begin{figure}[t]
\centering
    \subfigure[Sample discretized in a $128^3$ grid]
    {\label{fig:128_input}
    \includegraphics[width=.55\columnwidth]
    {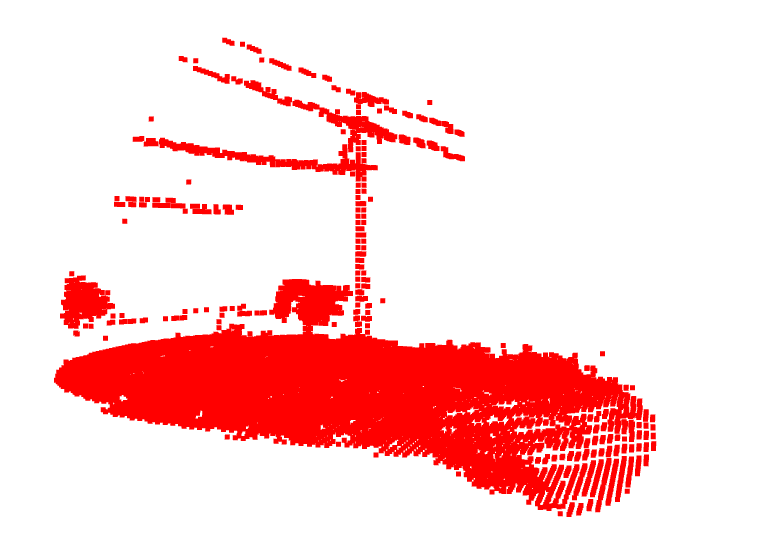}}
    \subfigure[SCENE-Net prediction against the GT]
    {\label{fig:128_pred}
    \includegraphics[width=.41\columnwidth]
    {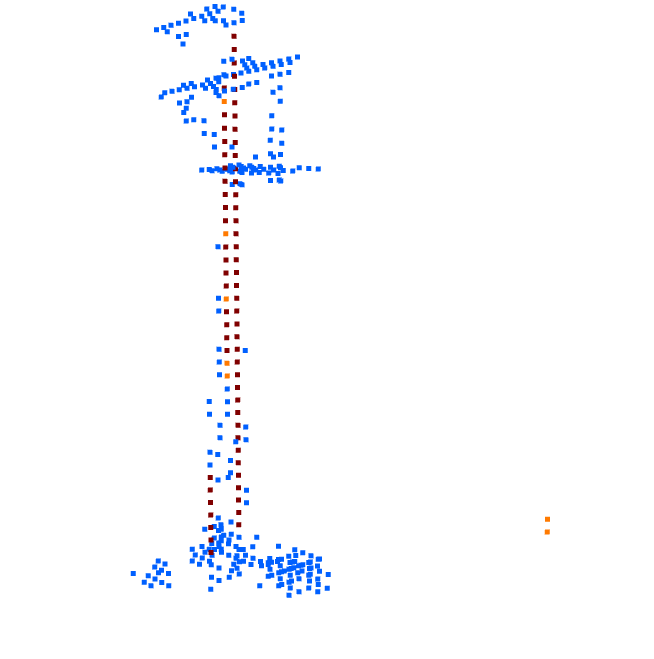}}
    \subfigure
    {\includegraphics[width=.5\columnwidth]
    {color_code_v2.pdf}}
    \caption{SCENE-Net is independent of the input and kernel size. Our model was trained with voxel grids of shape $64^3$ and kernel size $9^3$. Fig.~\ref{fig:128_pred} shows SCENE-Net's prediction against the ground truth of the $128^3$ input grid in Fig.~\ref{fig:128_input} using a kernel-size of $12\times5\times5$.  
}
\label{fig:vxg128}
\end{figure}

One of the issues of voxel-based models is the computational cost of  3D convolutions with large kernels and high-resolution voxel grids. Here, a CNN architecture leads to an exponential increase in training time.
SCENE-Net has a continuous functional observer of the raw input providing an analysis of its components.
Unlike traditional models, this definition is \textbf{independent} from the input size as well as its own discretization (kernel size). In this experiment, we trained SCENE-Net with voxel grids of $64^3$ and then applied to higher resolutions, such as $128^3$, with good qualitative results (Fig.~\ref{fig:vxg128}). 
The kernel size used to discretize the operators can be fine-tuned to enhance performance (Tab.~\ref{tab:kernel_sizes}).
%
%

\begin{table}[ht]
    \centering
    \caption{Performance of SCENE-Net with different kernel sizes on TS40K. SCENE-Net is trained with a kernel size of $9^3$, which is later fine-tuned to find the sweet spot between Precision and Recall.\\}
    \begin{tabular}{ccccc}
    \hline
    \multicolumn{2}{l}{Kernel-size}             & \multicolumn{1}{l}{Precision} & \multicolumn{1}{l}{Recall} & \multicolumn{1}{l}{IoU} \\ \hline
    \multicolumn{2}{c}{$9\times9\times9$}                &  0.37 ($\pm$ 0.02)                      & \textbf{0.22} ($\pm$ 0.01)                    & 0.58                  \\
    \multicolumn{2}{c}{$9\times5\times5$} & \textbf{0.68} ($\pm$ 0.08)                     & 0.16 ($\pm$ 0.05)                     & \textbf{0.58}                    \\ \hline
    \end{tabular}
    \label{tab:kernel_sizes}
\end{table}

\subsection{Template Matching Comparison.}

Template matching offers a direct approach to pattern detection by measuring the relation between the input data and a specific pattern of interest. 
Since we define the properties that constitute supporting towers, a comparison between SCENE-Net and this method is reasonable.

Our model combines GENEOs through convex combination and all shape parameters and convex coefficients are found through backpropagation, creating an aggregated GENEO operator - a data observer for composite shape signatures (e.g., cylinder+arrow+negative sphere).
In geometrical methods, such as template matching, this composition is not trivial.
Employing template matching on 3D data is slow, and the parameter initialization (e.g., a Cylinder radius) is a crucial step.
In addition, template matching is not directly applicable to some patterns, e.g., the Neg. Sphere, as it tries to diminish the activation of specific elements. 
We tested template matching of the TS40K with a Cylinder (the same definition used in SCENE-Net) with a radius equal to the average tower radius in the training set. It runs for 12 hours on the validation set of TS40K with an average precision (AP) of 0.001, while SCENE-Net’s AP = 0.2.

\subsection{Qualitative Results.}

\begin{figure*}[t]
\centering
\setkeys{Gin}{width=\textwidth}
    \addtocounter{subfigure}{-1}
    \subfigure{
    \includegraphics[width=.28\textwidth]
    {qualitative_res/s33/s33.png}}
    \addtocounter{subfigure}{-1}
    \subfigure{
    \includegraphics[width=.3\textwidth]{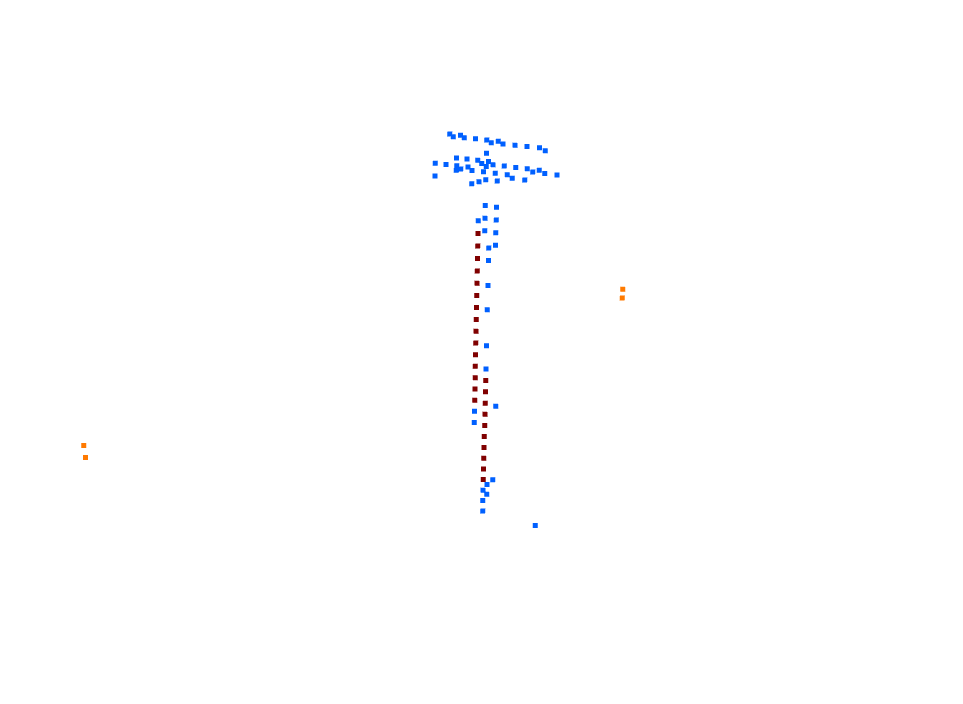}}
    \addtocounter{subfigure}{-1}
    \subfigure{
    \includegraphics[width=.3\textwidth]{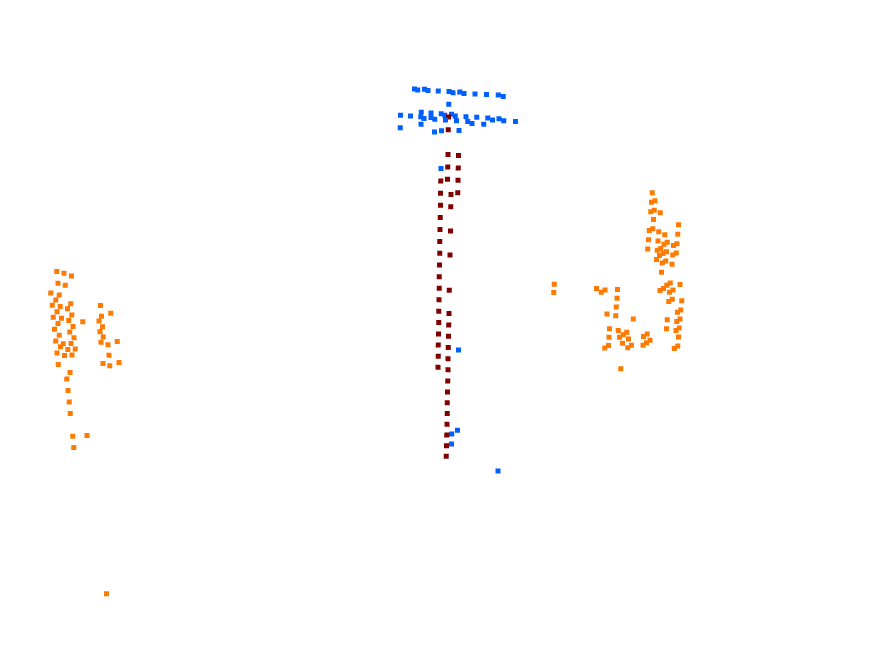}}
    
    \addtocounter{subfigure}{-1}
    \subfigure{
    \includegraphics[width=.28\textwidth]
    {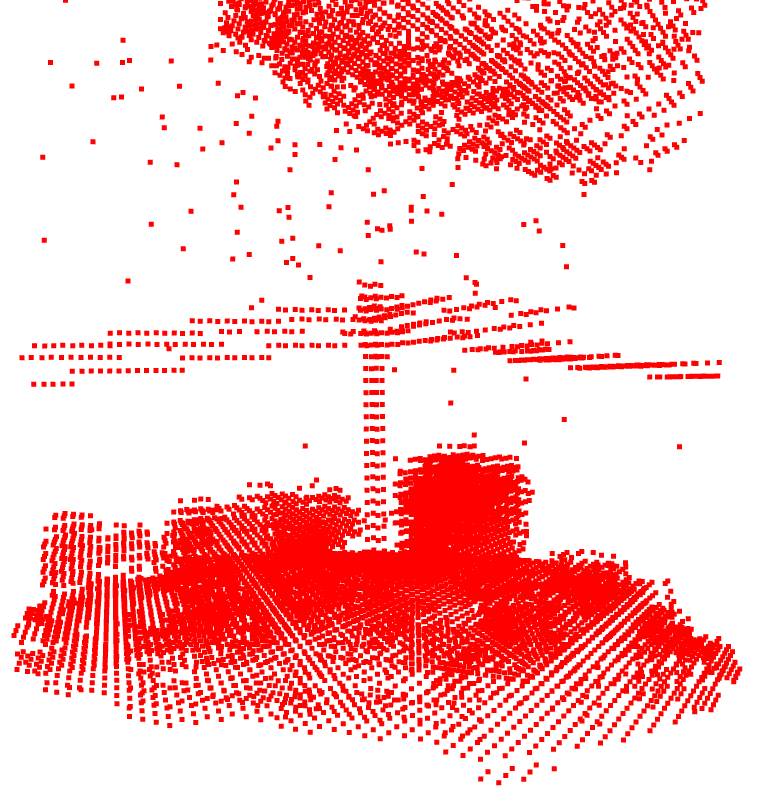}}
    \addtocounter{subfigure}{-1}
    \subfigure{
    \includegraphics[width=.3\textwidth]{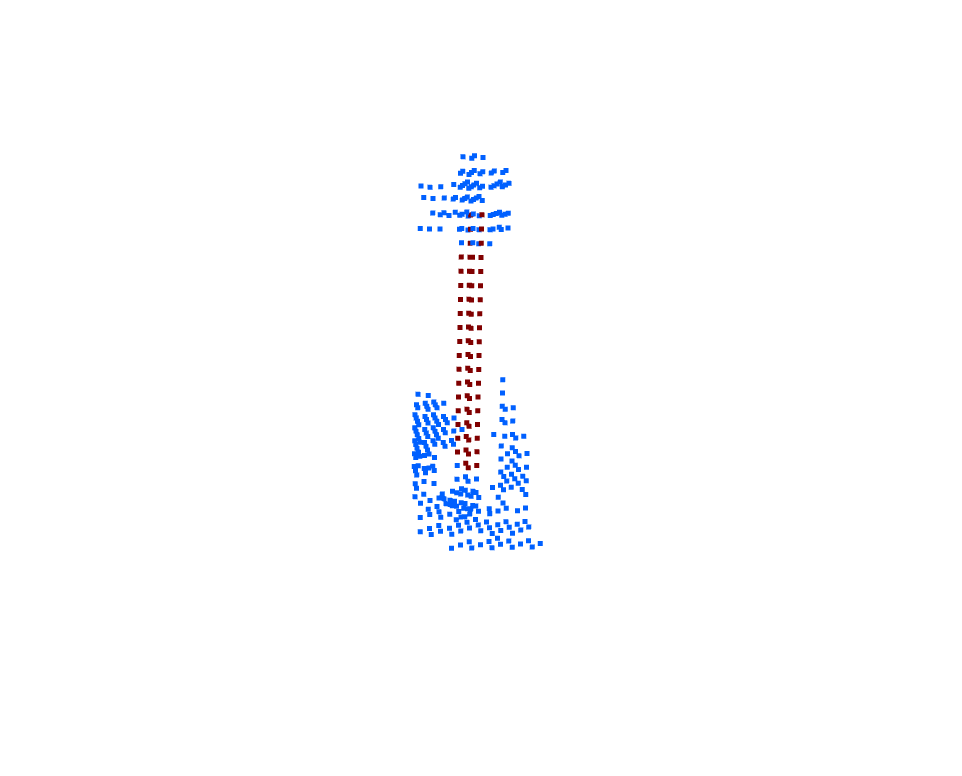}}
    \addtocounter{subfigure}{-1}
    \subfigure{
    \includegraphics[width=.3\textwidth]{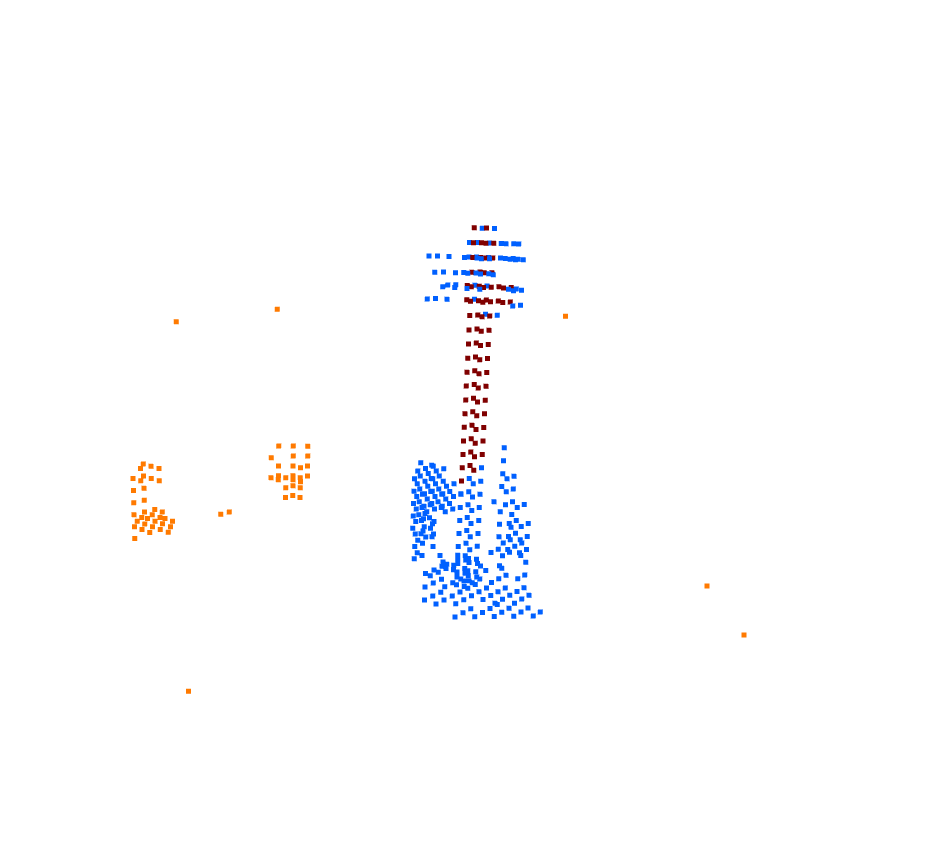}}

    \addtocounter{subfigure}{-1}
    \subfigure{
    \includegraphics[width=.28\textwidth]
    {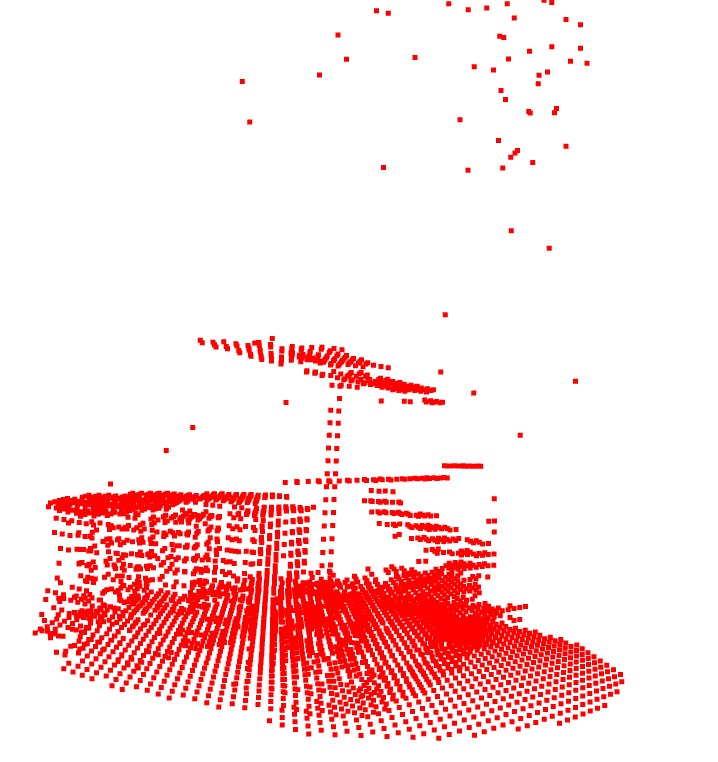}}
    \addtocounter{subfigure}{-1}
    \subfigure{
    \includegraphics[width=.3\textwidth]{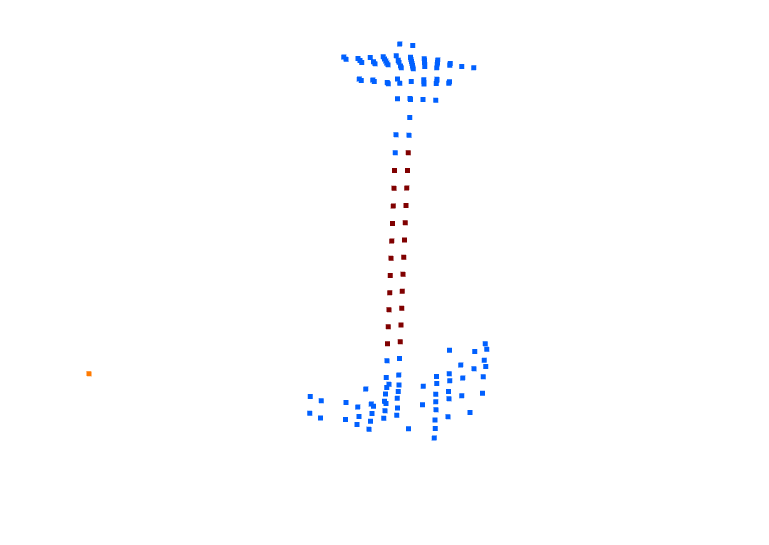}}
    \addtocounter{subfigure}{-1}
    \subfigure{
    \includegraphics[width=.3\textwidth]{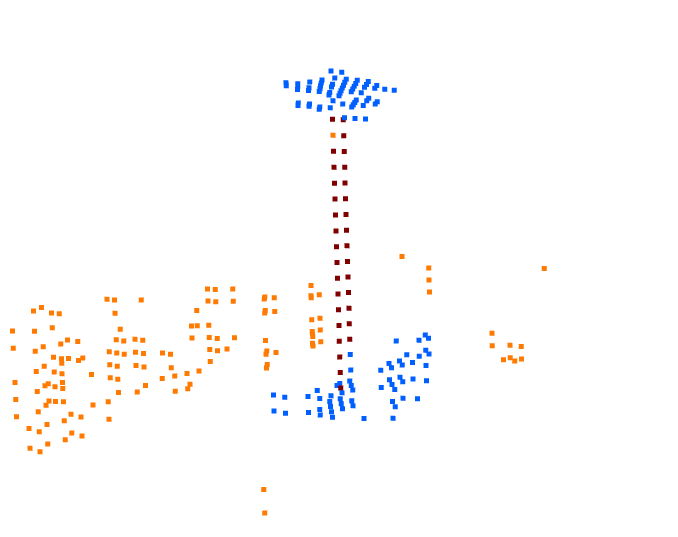}}

    \subfigure[TS40K Sample]{
    \includegraphics[width=.3\textwidth]
    {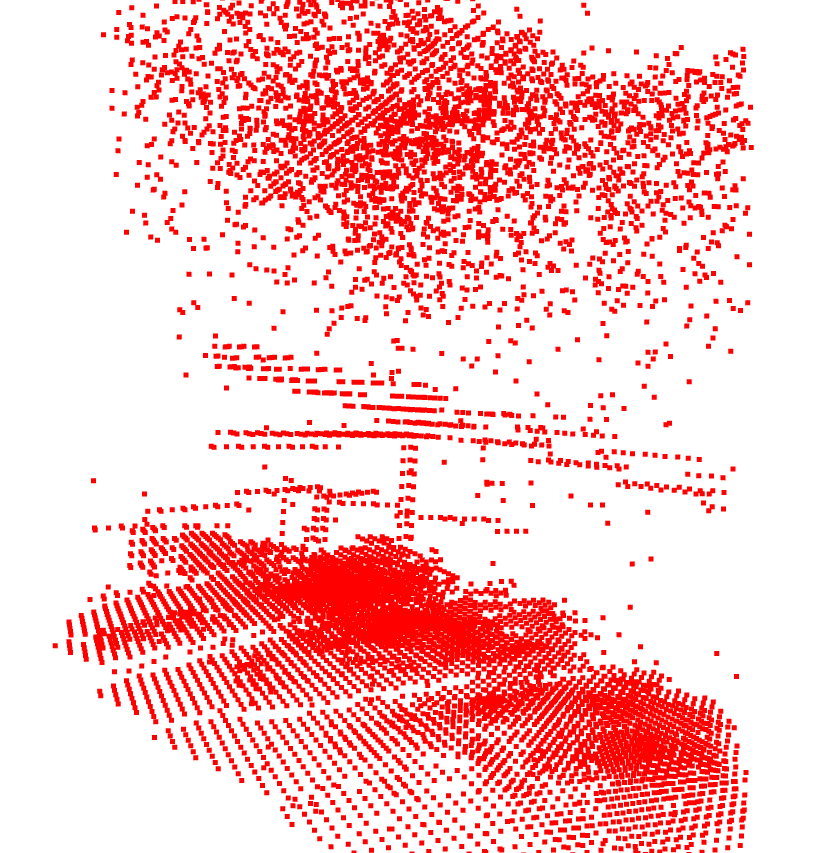}}
    \subfigure[SCENE-Net]{
    \includegraphics[width=.3\textwidth]{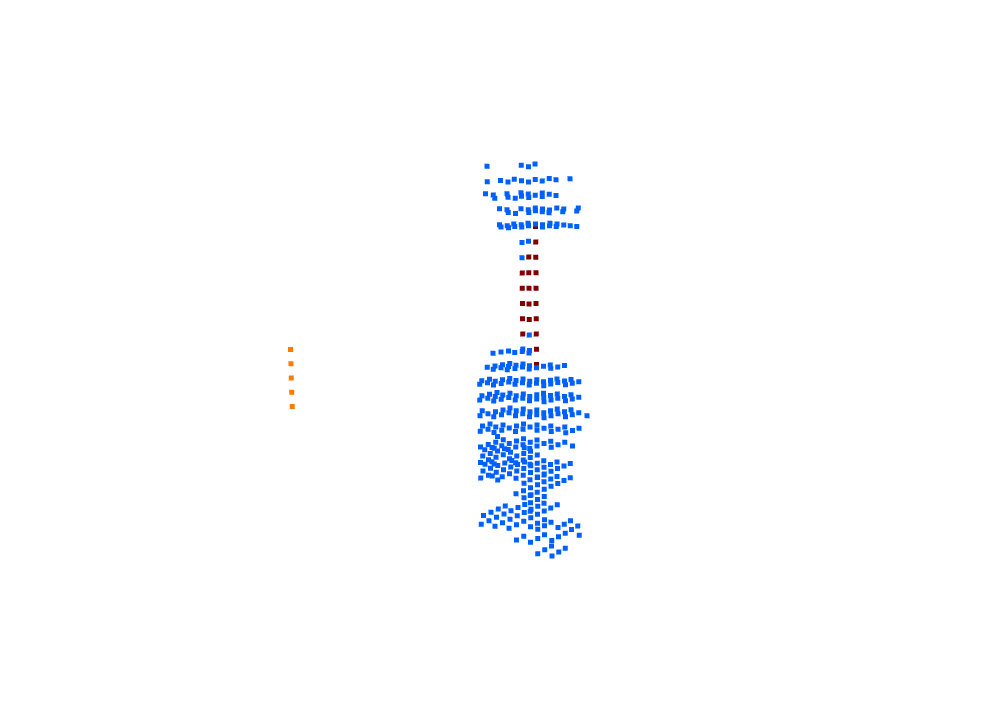}}
    \subfigure[CNN Baseline]{
    \includegraphics[width=.3\textwidth]{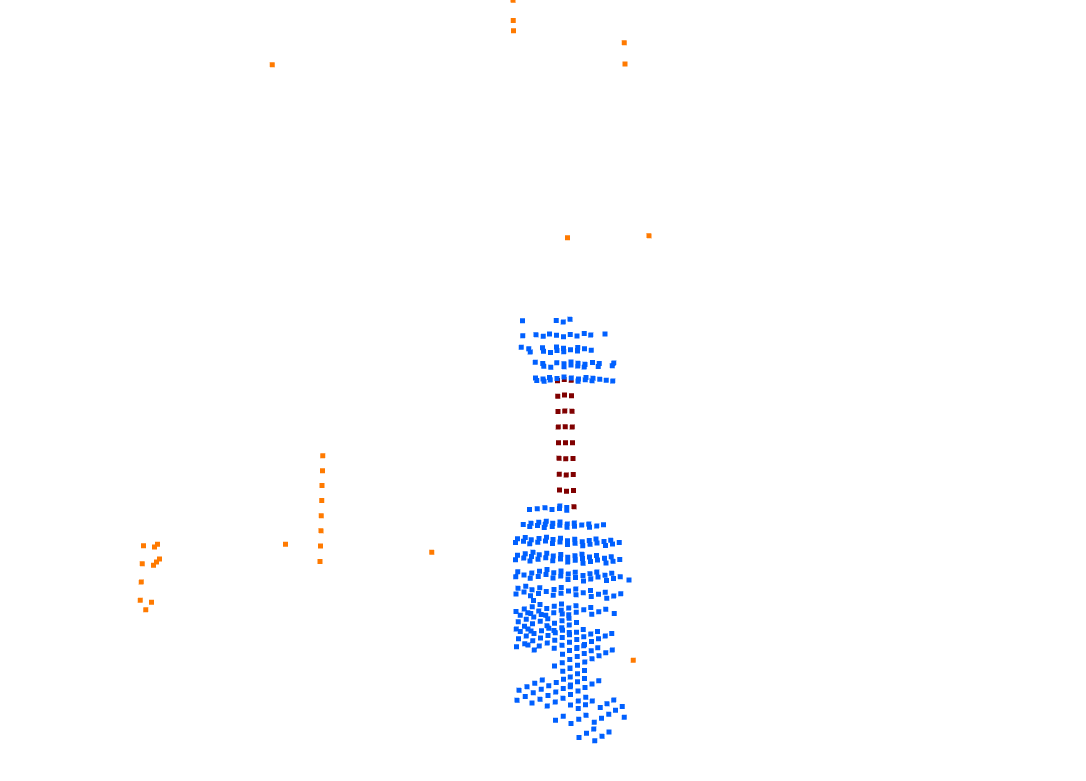}}

    \subfigure{
    \includegraphics[width=.5\columnwidth]
    {color_code_v2.pdf}}

    \caption{Qualitative results of SCENE-Net on the testing set of TS40K, against a CNN with similar architecture. Note that, in the last example, SCENE-Net clearly identifies a second unlabeled tower, whereas CNN identifies both the second tower and vegetation as towers.}
\label{fig:results}
\end{figure*}

In Figure~\ref{fig:results}, we can analyze the qualitative results on the TS40K dataset~(a) of SCENE-Net~(c) against a CNN is similar architecture as baseline~(b).
Even though the CNN achieves a higher Recall on the majority of the samples, it classifies most vertical scene elements as towers, which ultimately leads to a poor segmentation of supporting towers.
In contrast, SCENE-Net segments the body of towers and rejects other vertical objects that do not exhibit the prior knowledge encoded in the model.

\end{document}